\pdfoutput=1

\documentclass[11pt]{article}

\usepackage{acl}

\usepackage{times}
\usepackage{latexsym}

\usepackage[T1]{fontenc}

\usepackage[utf8]{inputenc}

\usepackage{microtype}

\usepackage{inconsolata}

\usepackage[disable]{todonotes}

\usepackage{tikz, tikz-qtree}
\usepackage{tikz-dependency}

\newcommand{\nl}[1]{\textit{{#1}}} %
\newcommand{\tbf}[1]{\textbf{{#1}}} %

\newcommand*{\ourdata}{\texttt{SPUD} }
\newcommand*{\ourdataNoSpace}{\texttt{SPUD}}

\newcommand*{\relacc}{RelAcc }
\newcommand*{\relaccNoSpace}{RelAcc}
\newcommand*{\uas}{UAS }
\newcommand*{\uasNoSpace}{UAS}
\newcommand*{\las}{LAS }
\newcommand*{\lasNoSpace}{LAS}

\usepackage[normalem]{ulem}

\usepackage{subcaption}

\usepackage{tabularx,booktabs}
\usepackage{multirow}

\usepackage{amssymb}

\title{
Multilingual Nonce Dependency Treebanks: Understanding how Language Models \textit{Represent} and \textit{Process} Syntactic Structure 
}

\author{David Arps$^{1}$, %
Laura Kallmeyer$^{1}$, %
Younes Samih$^{2}$, %
Hassan Sajjad$^{3}$\\
  Heinrich-Heine-Universität Düsseldorf$^{1}$, IBM Research$^{2}$, Dalhousie University$^{3}$\\
  \texttt{\{david.arps,laura.kallmeyer\}@hhu.de}, \texttt{younes.samih@ibm.com}, \texttt{hsajjad@dal.ca} \\
  }

\begin{document}
\maketitle

\begin{abstract}
We introduce \texttt{SPUD} (Semantically Perturbed Universal Dependencies), a framework for creating nonce treebanks for the multilingual Universal Dependencies (UD) corpora. \texttt{SPUD} data satisfies syntactic argument structure, provides syntactic annotations, and ensures grammaticality via language-specific rules. We create nonce data in Arabic, English, French, German, and Russian, and demonstrate two use cases of SPUD treebanks. 
First, we investigate the effect of nonce data on word co-occurrence statistics, as measured by perplexity scores of autoregressive (ALM) and masked language models (MLM). We find that ALM scores are significantly more affected by nonce data than MLM scores. Second, we show how nonce data affects the performance of syntactic dependency probes. We replicate the findings of \citet{muller-eberstein-etal-2022-probing} on nonce test data and show that the performance declines on both MLMs and ALMs wrt.~original test data. However, a majority of the performance is kept, suggesting that the probe indeed learns syntax independently from semantics.\footnote{Code at \url{https://github.com/davidarps/spud}}%
\end{abstract}

\section{Introduction}\label{sec:intro}
\todo{Reviewer: Both the presentation and conclusion end up being parceled and unfocused. Better now?}
\todo{Adress Review: Mention Faithful Interpretability}

An ample amount of work in the last years has focused on making explicit the linguistic information encoded in language models (LMs). %
Frequently, the overarching question is to which extent the behavioral and representational properties of self-supervised LMs are cognitively plausible and assumed by linguistic theory~\cite[surveys:][]{linzen-baroni-2021-syntactic,mahowald-etal-2023-dissociating,chang-bergen-2023-language}. %
A subset of this work aimed at understanding LMs ability to learn syntax (survey: \citealp{kulmizev-nivre-2022-schrodingers}, examples: \citealp{hewitt-manning-2019-structural,manning-etal-2020-emergent,tenney-etal-2019-bert,newman-etal-2021-refining}). 
A common approach %
 is to rely on the performance of a probing classifier \cite{hupkes-etal-2018-visualisation,kunz-kuhlmann-2020-classifier,kunz-kuhlmann-2021-test}
 in predicting syntactic relations. 
 However, this method has been criticized for various reasons \cite{belinkov-2022-probing}, including that predicting a linguistic property from LM representations does not imply that the LM uses that property in its predictions \cite{lyu-etal-2023-faithful}. 
\citet{hall-maudslay-cotterell-2021-syntactic} highlighted that results of syntactic probes %
 can be misleading due to semantic information in the representation. 
In other words, a high probing accuracy may be fully or partially attributed to the use of present semantic knowledge. 

The theoretical linguistic literature has discussed the separation of syntactic grammaticality and semantic information for decades, starting with the sentence \textit{Colorless green ideas sleep furiously} in \citet{chomsky-1957-syntactic-structures}. 
A number of studies have used systematically perturbed data to probe LM representations and predictions on such grammatical but nonsensical sentences~\cite{gulordava-etal-2018-colorless,hall-maudslay-cotterell-2021-syntactic,arps-etal-2022-probing}. 
 However, none of the works ensures that the nonce data is grammatical, %
 e.g. in their datasets, the valency of predicates is not controlled. 
Furthermore, only \citet{gulordava-etal-2018-colorless} base their conclusions on languages beyond English.

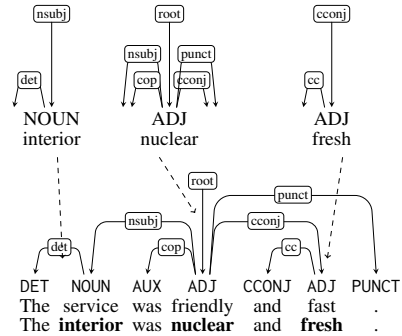
\begin{figure}\centering
    \centering\scalebox{.67}{
      \begin{tikzpicture}
      \node at (0cm, 0cm) {
      \begin{dependency}
      \begin{deptext}
          \texttt{DET} \& \texttt{NOUN} \& \texttt{AUX} \& \texttt{ADJ} \& \texttt{CCONJ} \& \texttt{ADJ} \& \texttt{PUNCT} \\
          The \& service \& was \& friendly \& and \& fast \& . \\
          The \& \tbf{interior} \& was \& \tbf{nuclear} \& and \& \tbf{fresh} \& . \\
      \end{deptext}
      \depedge{2}{1}{det}
      \deproot{4}{root}
      \depedge{4}{2}{nsubj}
      \depedge{4}{3}{cop}
      \depedge{4}{6}{cconj}
      \depedge{4}{7}{punct}
      \depedge{6}{5}{cc}
  \end{dependency}};
  
  \node at (-32mm, 35mm) {
      \begin{dependency}
          \begin{deptext}
              \& NOUN \\
              \& interior \\
          \end{deptext}
          \deproot{2}{nsubj}
          \depedge{2}{1}{det}
      \end{dependency}};
  
  \node at (-8mm, 35mm) {
      \begin{dependency}
          \begin{deptext}
              \& \& ADJ \& \& \\
              \& \& nuclear \& \& \\
          \end{deptext}
          \deproot{3}{root}
          \depedge{3}{1}{nsubj}
          \depedge{3}{2}{cop}
          \depedge{3}{4}{cconj}
          \depedge{3}{5}{punct}
      \end{dependency}};
  \node at (23mm, 35mm) {
      \begin{dependency}
          \begin{deptext}
              \& ADJ \\
              \& fresh \\
          \end{deptext}
          \deproot{2}{cconj}
          \depedge{2}{1}{cc}
      \end{dependency}};
  
  \draw[->, dashed] (-30mm, 20mm) -- (-29mm, 0mm);
  \draw[->, dashed] (-10mm, 20mm) -- (-3mm, 9mm);
  \draw[->, dashed] (26mm, 20mm) -- (23mm, 0mm);
  \end{tikzpicture}
  }
      \caption{\ourdata data creation}\label{fig:nonce-data-general-algo-example}
\end{figure}

In this paper, we aim to standardize the efforts in this direction by proposing a framework to create nonce data that satisfies syntactic argument structure, provides syntactic annotations, and ensures grammaticality via language-specific rules. The framework relies on Universal Dependencies \cite[UD,][]{de-marneffe-etal-2021-universal} treebanks. For each sentence in the treebank, it replaces content words with other content words that are known to appear in the same syntactic context. This results in nonce sentences that preserve the %
 syntactic structure of %
  the original sentences (see Fig.~\ref{fig:nonce-data-general-algo-example}). To ensure syntactic grammaticality, we define three types of language-specific rules and constraints that are applied during the process, which are:
(i) the POS tag of the words to be replaced, 
(ii) constraints on word order and dependency relations of replaced words to ensure syntactic grammaticality, and
(iii) word-level rules to ensure that morphosyntactic constraints are met. 
We refer to this algorithm and the resulting UD data as \emph{Semantically Perturbed UD} (\ourdataNoSpace).

We create \ourdata data for five widely spoken
languages; Arabic, English, French, German, and Russian. 
We show via a human evlation that \ourdata is preferrable in terms of grammaticality to the kind of nonce data that has been previously used in the literature to tackle similar research questions. 
We show the effectiveness of \ourdata
on two tasks to assess the robustness of autoregressive LMs (ALMs) and masked LMs (MLMs) to semantic perturbations. 
First, we study the effect of nonce data on LM scoring functions (perplexity and its adaptations for MLMs). 
Second, we investigate the robustness of syntactic dependency probes to semantic irregularities, and disentangle the effect of lexical semantic features on the findings of previous work. 
The contributions and main findings of our work are as follows:
\begin{itemize}
\setlength\itemsep{0em}
    \item We introduce \ourdataNoSpace, a framework for creating nonce treebanks for UD %
     corpora that ensures syntactic grammaticality, and provide nonce treebanks for 5 languages (Arabic, English, French, German, Russian).
    \item We show the effectiveness of the proposed data log-likelihood scoring of different language model architectures, and on the performance of syntactic dependency probes.
    \item We show that ALM perplexity is significantly more affected by nonce data than two formulations of MLM pseudo-perplexity, and that for MLMs, the availability of subword information affects pseudo-perplexity scores on both natural and \ourdata data. 
    \item In structural probing for dependency trees, we show that ALM performance decreases more than MLM performance on \ourdataNoSpace, and that this performance drop is more pronounced for edge attachment than relation labeling. 
\end{itemize}

The paper is structured as follows. 
In the next section, we %
 discuss %
  related work. %
In Sec.~\ref{sec:nonce-data}, we describe the framework for creating nonce UD treebanks. %
Sec.~\ref{sec:perplex} describes the experiment on scoring nonce data with perplexity.  
In Sec.~\ref{sec:nonce-probing}, we describe the structural probing experiments. 
In Sec.~\ref{sec:conclusions}, we discuss the results and conclude.

\section{Related Work}\label{sec:related-work}

\paragraph{Automatically modified Dependency Trees} 

The idea to replace words and subtrees based on dependency structure has been used for %
 data augmentation %
  in various areas of NLP, in particular for 
  low-resource dependency parsing. 
\citet{dehouck-gomez-rodriguez-2020-data} proposed an idea similar to ours, with the differences that (i) they swap subtrees instead of content words resulting in generated sentences with altered syntactic structure, and 
(ii) their constraints on possible replacements are language-agnostic. 
\citet{sahin-steedman-2018-data} cut subtrees based on their dependency relation, and modified treebanks by changing the order of annotated dependency subtrees.
\citet{vania-etal-2019-systematic} demonstrated the effectiveness of \citet{sahin-steedman-2018-data}'s algorithm as well as nonce treebanks for low-resource dependency parsing. However, their method 
lacked language-specific constraints and did not rely on dependency edges to dependents.
\citet{nagy-etal-2023-data} apply a similar 
method 
for low-resource machine translation. 
\citet{wang-eisner-2016-galactic} permuted dependents within a single tree to generate synthetic treebanks with lexical contents from one language and word order properties of another language.

\paragraph{Structural Probing}\label{sec:structural-probes}

DepProbe \cite{muller-eberstein-etal-2022-probing} decodes labeled and directed dependency trees from LM representations. 
With DepProbe, \citet{muller-eberstein-etal-2022-probing} probed mBERT \cite{devlin-etal-2019-bert} on 13 languages, and showed that the probe performance is predictive of parsing performance after finetuning the LM. 
Prior to that, similar probes have been applied to unlabeled \cite{kulmizev-etal-2020-neural} as well as unlabeled and undirected dependency trees \cite{hewitt-manning-2019-structural,chi-etal-2020-finding}. %
\citet{eisape-etal-2022-probing} used \citet{hewitt-manning-2019-structural}'s method to probe GPT-2 \cite{radford-etal-2019-gpt2}. 

\paragraph{Syntactic and Semantic Information in LMs}
A number of works interpret deep learning models for syntax and semantics at representation-level~\cite{belinkov:2017:acl,blevins-etal-2018-deep} and at neuron level~\cite{sajjad-etal-2022-neuron,durrani_neuron_jmlr_2023,torroba-hennigen-etal-2020-intrinsic}. 
\citet{gulordava-etal-2018-colorless} performed targeted syntactic evaluation \cite[TSE;][]{marvin-linzen-2018-targeted} on ALMs %
 in four languages and investigated the effect of nonce data that considers POS tags and morphological features for replacements. 
\citet{lasri-etal-2022-bert} investigated the effect of nonce data for TSE in BERT and found that BERT correctly predicts number agreement for nonce sentences only in simple syntactic templates. 
\citet{hall-maudslay-cotterell-2021-syntactic} investigated the effect of \textit{Jabberwocky words} (such as \textit{provicated}) on syntactic probes. 
\citet{ravfogel-etal-2020-unsupervised} found a transformation of LM representations that
highlights structural properties. 
\citet{arps-etal-2022-probing} used a similar algorithm as ours to create nonce versions of the English PTB \cite{marcus-etal-1993-building}, and probed the syntactic information in hidden representations of four MLMs. 
The main difference to %
\ourdata
is that they do not restrict replacements to content words, and do not apply language-specific processing steps. Therefore their nonce data is more likely to be ungrammatical. %
\citet{sinha-etal-2021-masked} tested to which extent MLMs rely on word order %
vs. higher-order cooccurence statistics. %
They found that word order is not needed to achieve high performance on many NLP tasks. %
\citet{papadimitriou-etal-2022-classifying-grammatical} probed the relevance of word order and semantic prototypicality for classifying grammatical roles with BERT. 
\citet{lasri-etal-2022-subject} compared BERT and human judgments on the subject-verb agreement task for nonce data and found that, while the error patterns of both are similar, BERT has a generally higher performance drop for nonce data than humans.
\citet{kauf-etal-2023-lexical} explored the interaction of syntactic and semantic information with similarity between LMs and human fMRI signals, and show that lexical semantic content - not syntactic structure - is the main driver of similarity in both LM and human representations. 

\paragraph{Scoring Functions for LMs}
The likelihood of sentences assigned by an LM is often used to investigate the model's preference for grammatical sentences~\cite{kulmizev-nivre-2022-schrodingers,warstadt-etal-2020-blimp}. 
Perplexity ($PPL$) is a common metric to score the likelihood of a sentence with ALMs. For MLMs, estimating the likelihood of a given sentence in a fashion that is useful for applications is not as trivial. 
\citet{salazar-etal-2020-masked} formalized pseudo-perplexity ($PPPL$) as a scoring function for MLMs: %
Each token in a sentence is masked in turn, and $PPPL$ is computed as a function of the MLM's probability of the masked tokens.
\citet{kauf-ivanova-2023-better} showed that $PPPL$ systematically inflates scores for multi-token words, and proposed $PPPL_{l2r}$, a subword-aware alternative that is more robust to this effect. 
\citet{miaschi-etal-2021-makes} trained classifiers on different linguistic features to predict the $PPL$ of GPT-2 and $PPPL$ of BERT. 
They find that GPT-2 scores are generally better predictable by their set of features than BERT scores, and that lexical features are more important for GPT-2 than for BERT.

\section{Nonce Treebanks for Five Languages}\label{sec:nonce-data}
This section presents \ourdata (\textit{Semantically Perturbed Universal Dependencies}), our framework for nonce data creation. 
The input %
 is a UD treebank, and the output is a treebank with the same syntactic structure as the input but nonce semantic content. %
 In principle, the same algorithm is applied in any language, but each language requires a set of language-specific pre- and post-processing steps (Sec.~\ref{sec:alg-lang-independent}). 
In this work, we create nonce data for five languages and treebanks: 
Arabic \cite[PADT,][]{hajic-etal-2009-praguearabic}, 
German \cite[HDT,][]{borges-volker-etal-2019-hdt}, 
English \cite[EWT,][]{silveira-etal-2014-gold}, 
French \cite[GSD,][]{guillaume-etal-2019-conversion}, 
and Russian \cite[SynTagRus,][]{droganova-etal-2018-data}. 
Fig.~\ref{fig:nonce-data-general-algo-example} presents one example. 
More examples and information on the resources are in App.~\ref{app:nonce}. 
To apply the framework to a new language, an annotated UD treebank and access to a native speaker in the target language are required. Along with this paper, we publish a tutorial for creating %
 \ourdata treebanks in other languages.

\subsection{Generating Nonce Data}\label{sec:alg-lang-independent}

\paragraph{Language-independent algorithm}
The procedure consists of iteratively replacing content words with other words that appear in the same syntactic context at another point in the treebank. 
We consider words with POS tags \texttt{ADJ}, \texttt{ADV}, \texttt{NOUN}, \texttt{PROPN} and \texttt{VERB} as content words. The syntactic context of a token $t$ is defined as 
(i) the UPOS tag of $t$, 
(ii) the dependency relation of $t$ to its head, and 
(iii) the dependency relations of $t$ to its dependents. 
This syntactic context is collected for every lemma. 
Then, the nonce trees are created by replacing content words with other words where the lemma appeared in the same syntactic context. 
The morphological features of $t$, as annotated in UD, are considered to determine the right form for a replacement. For this step, the morphological databases UDLexicon \citep{sagot-2018-multilingual} and Wiktextract \cite{ylonen-2022-wiktextract} are used.  
Fig.~\ref{fig:nonce-data-general-algo-example} presents an example (without morphosyntactic features).

\paragraph{Language-specific modifications}\label{sec:alg-lang-specific}
 are necessary to ensure that the data meets the criterion of being morphosyntactically correct. 
For instance, if the first sound of the word following an English indefinite article changes from a vowel to a consonant or vice versa, 
 the article is adjusted with the help of a phonological dictionary. 
 E.g. when replacing \nl{apple} in \nl{an apple} with \nl{bicyle}, the result is \nl{a bicycle} and not \nl{*an bicycle}.
We refer to App.~\ref{app:nonce} 
for details on the language-specific modifications.

\paragraph{Quality of the generated data}%
To determine the sufficiency of the language specific rules, we asked linguistically trained native speakers of the respective languages to provide feedback on how well the nonce sentences match the desired criteria. 
The annotators received samples of at least 100 sentence pairs. 
Over 2-3 iterations, annotators pointed out problems in the generated data, which we then addressed by modifying the language-specific rules. 

\paragraph{Human evaluation}
\todo{Say that we as the authors annotated, or keep it vague?}
We conduct a human evaluation to estimate the benefits of using SPUD over the algorithm presented by \citet{gulordava-etal-2018-colorless}, which does not incorporate information about syntactic dependencies or language-specific rules. 
For this evaluation, human annotators were presented with sentence triplets, each consisting of an original sentence from the treebank, a nonce version of that sentence generated by SPUD, and a nonce version generated by the algorithm of \citet{gulordava-etal-2018-colorless}. 
The two nonce sentences were presented in random order, without indication of their source. 
Annotators rated for each sentence whether it was grammatical, and which of the two nonce sentences was syntactically closer to the original. 
One annotator rated 30 French sentence triplets, and two annotators rated 39 English and 153 German sentence triplets. 
All triplets where selected randomly from the corresponding treebanks. 
Annotators had the option to indicate that they were unsure, or that the two nonce sentences were equally close to the original. 
We find that SPUD is rated grammatical more frequently than the algorithm of \citet{gulordava-etal-2018-colorless} in all three languages, and by all annotators (Tab.~\ref{tab:judgments-humaneval}). 
Concretely, up to 87\% of the SPUD sentences are rated grammatical, compared to only up to 38\% of the sentences generated by the algorithm from \citet{gulordava-etal-2018-colorless}. 
Inter-annotator agreement on grammaticality judgments is moderate, with Cohen's Kappa of .51 (en) and 0.58 (de). 
SPUD is also preferred in the majority of cases, with some "equal" ratings (Tab.~\ref{tab:preferenes-humaneval}). 
We conclude that, while the scale of this experiment is limited, the results suggest that the quality of the sentences generated by SPUD is higher than that of the algorithm of \citet{gulordava-etal-2018-colorless}, which has been used previously to interpret the behavior of LMs. 

\begin{table*}
    \centering
    \begin{tabular}{lllllllll}
        \toprule
        {} & & \multicolumn{3}{c}{SPUD}  &  \multicolumn{3}{c}{Gulordava et al. (2018)}  &  \\
        \midrule
            &  Sentences &   yes &         no &     unsure &                     yes &         no &     unsure &     $\kappa$ \\
        de  &                153 &  72.5 &       27.5 &          0 &                     9.5 &       90.5 &          0 &       0.58 \\
        en  &                 39 &  78.2 &       19.2 &        2.6 &                    35.9 &       60.3 &        3.8 &       0.51 \\
        fr  &                 30 &  86.7 &       13.3 &          0 &                       0 &        100 &          0 &            \\
        \bottomrule
        \end{tabular}
    \caption{Results for Grammaticality judgments. For German and English, the mean between the two annotators is presented. 
    Grammaticality judgments were also made for the original sentences, which were almost always rated grammatical. They are not included in the figure and in the inter-annotator agreement calculations.}
    \label{tab:judgments-humaneval}
    \end{table*}
    
    \begin{table*}
    \centering
    \begin{tabular}{lrrrrr}
        \toprule
        {} &  Equal &  SPUD &  Gulordava et al. (2018) &  unsure &  $\kappa$ \\
        \midrule
        de &   13.7 &  81.0 &                      4.2 &     1.0 &    0.17 \\
        en &   20.5 &  70.5 &                      1.3 &     7.7 &    0.09 \\
        fr &   13.3 &  86.7 &                      0.0 &     0.0 &         \\
        \bottomrule
        \end{tabular}
    \caption{Results for Preference judgments. For German and English, the mean between the two annotators is presented.
    While Cohen's $\kappa$ is low, per item agreement is higher (.56 for en, .73 for de).}
    \label{tab:preferenes-humaneval}
    \end{table*}

\section{Scoring \ourdata with ALMs and MLMs}\label{sec:perplex}

The \ourdata treebanks %
 are designed to be grammatical but highly improbable. 
In this section, we investigate how this %
 property is reflected in the predictions of different LM architectures. 
On the one hand, this serves as a sanity check for %
 \ourdata resources: 
Do we in fact perturb co-occurrence statistics of words as intended? On the other hand, at a higher level, it investigates whether \ourdata data makes the models perplexed and how different LM architectures, and uni- vs. bidirectional context, influence the effect that syntactic and semantic structure has on model predictions. 
Concretely, we answer the following questions:
To what extent is \ourdata data harder to predict than original data? Are ALMs and MLMs affected by \ourdata data in different ways? Are MLM scoring functions affected differently by nonce data?

\subsection{Scoring Functions for LMs} 

App.~\ref{app:scoring-ex} shows examples for all scoring functions.

\paragraph{ALMs: Perplexity ($PPL$)} 
of a sentence $x = (x_1,\dots,x_n)$ is commonly defined as the exponentiated average of the negative sum of
log probabilities for all tokens $x_i$. The lower the perplexity, the higher
  the probability of $x$ %
   for the model. 
\begin{equation}\label{eq:ppl}
  PPL(x) = exp(-\frac{1}{n} \sum_{i=0}^n \log p(x_i|x_{<i}))
\end{equation}

\paragraph{MLMs: Pseudo-Perplexity ($PPPL$)} 
is designed to capture the likelihood that an MLM assigns to a sequence~\citep{salazar-etal-2020-masked}. 
It is calculated by processing 
 the input %
 $n$ times, masking each token $x_i$ exactly once. 
 $PPPL(x)$ %
  is 
  defined via the sum of log probabilities that the LM assigns to the masked %
 $x_i$: 
\begin{equation}\label{eq:pppl}
    PPPL(x) = exp(-\frac{1}{n} \sum_{i=0}^n \log p(x_i|x_{\setminus i}))
\end{equation}

\paragraph{MLMs: $PPPL$ with subword generation}
\citet{kauf-ivanova-2023-better} proposed $PPPL_{l2r}$, a variant of $PPPL$ that is more aligned with $PPL$ for ALMs. 
The idea is to consider the groupings of tokens to words, and predict subword tokens without conditioning on succeeding tokens. 
Let $\omega_r(x_i)$ denote the tokens that are in the same word as $x_i$, including and succeeding $x_i$. 
For example, when tokenizing \textit{accordeon} as \textit{accord, \#\#eon}, it holds that 
$\omega_r(\textrm{accord}) = \{\textrm{accord, \#\#eon}\}$. 
Then, tokens $\omega_r(x_i)$ from the right context within the same word are masked out when predicting the token $x_i$: 
\begin{equation}\label{eq:pppl-l2r}
    PPPL_{l2r}(x) = exp(-\frac{1}{n} \sum_{i=1}^{n} \log p(x_{i}|x_{\setminus \omega_r(x_i)}))
\end{equation}

\paragraph{Evaluating Scoring Functions on \ourdata }
When comparing scores across sentences, the variance is very high: 
Some sentences are far more likely than others, irrespective of the language or whether they are nonce. 
Thus, we use
the ratio between sentence-level scores of a pair of original and corresponding \ourdata sentence $(s_{orig}, s_{nonce})$. 
For a scoring function $f \in \{PPL, PPPL, PPPL_{l2r} \}$, we define the ratio as
$r_f(s_{orig}, s_{nonce}) = \frac{f(s_{nonce})}{f(s_{orig})}$
and then investigate the distribution of $r_f$ for all sentence pairs in a corpus.

\subsection{Hypotheses}
Hypothesis 1: \textit{LMs assign a higher score to \ourdata data than to original data} - the content words are chosen at random, and therefore should be much harder to predict than the original words. 

\noindent Hypothesis 2: \textit{ALMs and MLMs are impacted by \ourdata data in different ways. }
Bidirectional context determines the syntactic properties of a predicted token to a larger degree than unidirectional context. 
Therefore, the space of probable predictions is smaller for MLMs than for ALMs, and we expect that 
the impact of \ourdata data on ALM perplexity should be higher than on MLM pseudo-perplexity. 

\noindent Hypothesis 3: \textit{For \ourdataNoSpace , $r_{PPPL_{l2r}}$ are higher than $r_{PPPL}$} because nonce words are hard to predict based on context from surrounding words, but easy to predict based on context from the same word.

\subsection{Experimental Setup}

We adapt \citet{kauf-ivanova-2023-better}'s implementation to retrieve token-level scores for all sentences in the \ourdata and original data. 
We compare results for mBERT \cite{devlin-etal-2019-bert} as an MLM and mGPT \cite{shliazhko-etal-2022-mgpt} as an ALM.
We report the scoring results of only one instance of \ourdata data per sentence. 
We argue that this is sufficient to draw conclusions about the general behavior of the models, since we conduct the experiment on a large number of sentences, and different languages. 

\subsection{Results}
\begin{figure}
\centering
\scalebox{.6}{
\begin{tabular}{lccc}
\toprule
{} & $r_{PPPL}$ & $r_{PPPL_{l2r}}$ & $r_{PPL}$ \\
\midrule
ar & \includegraphics[width=.2\textwidth]{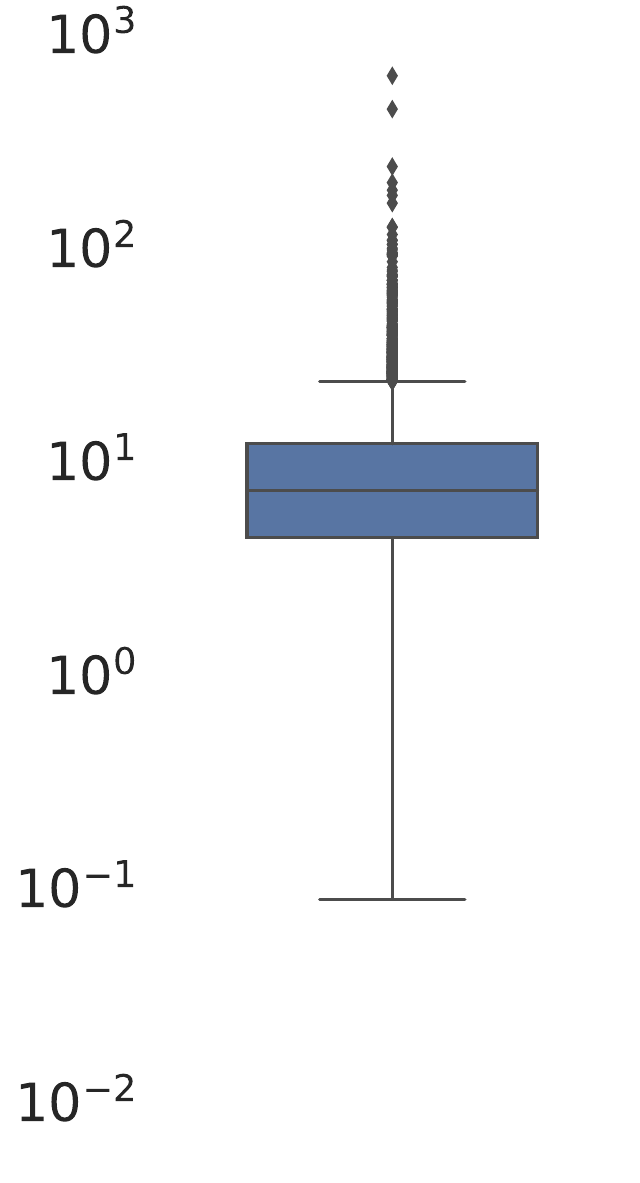} & \includegraphics[width=.2\textwidth]{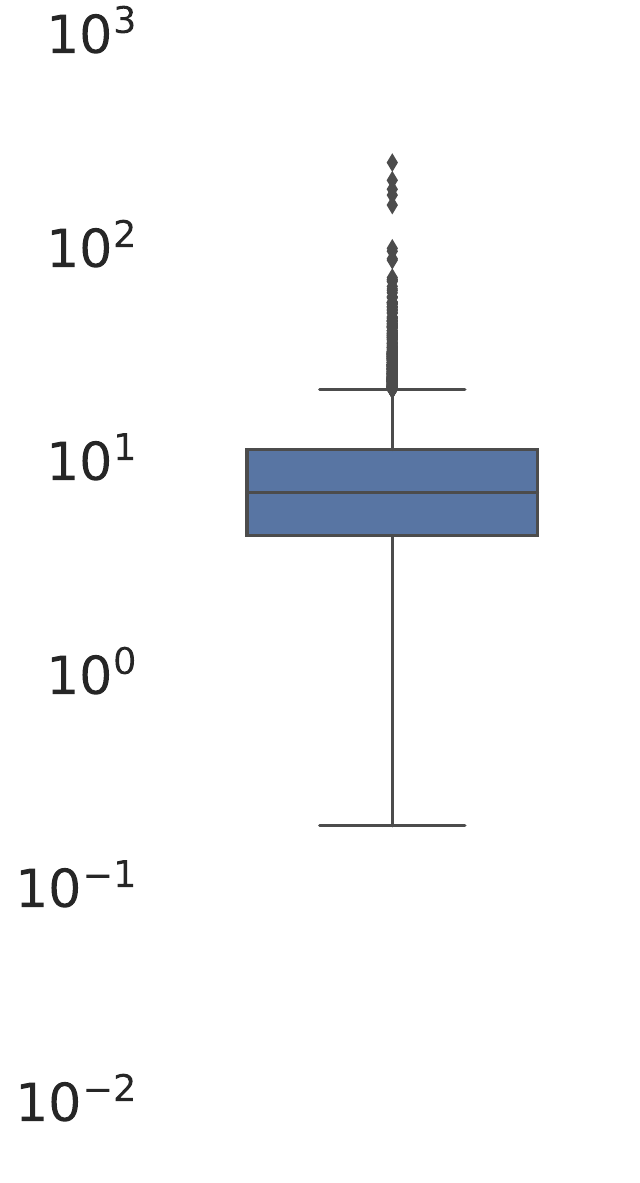} & \includegraphics[width=.2\textwidth]{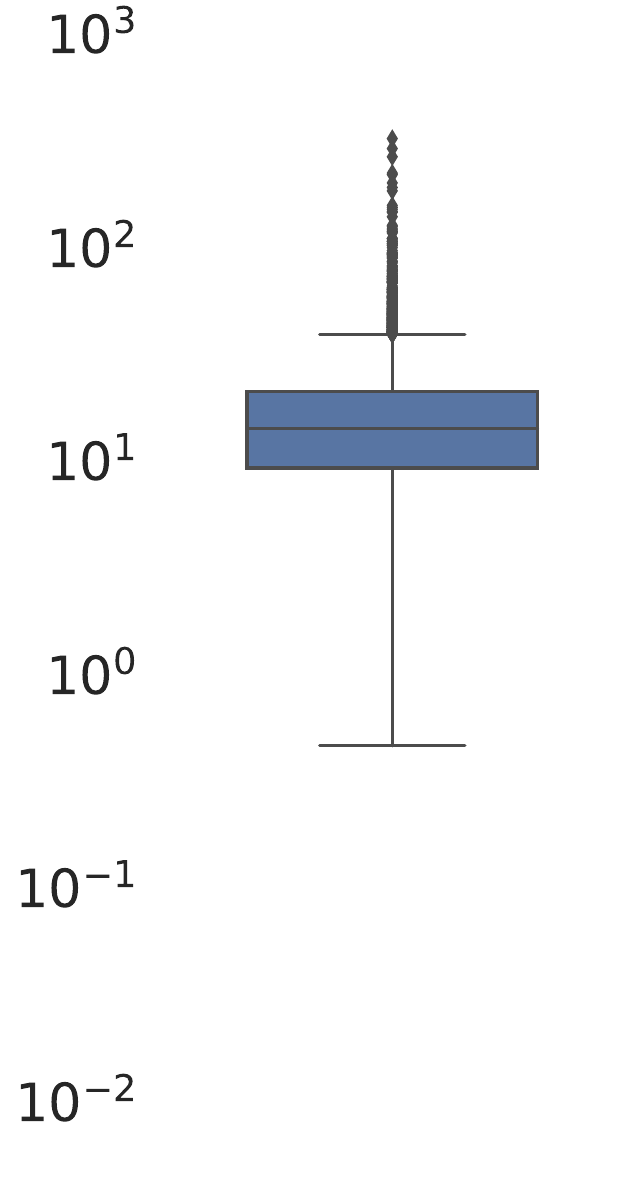} \\
en & \includegraphics[width=.2\textwidth]{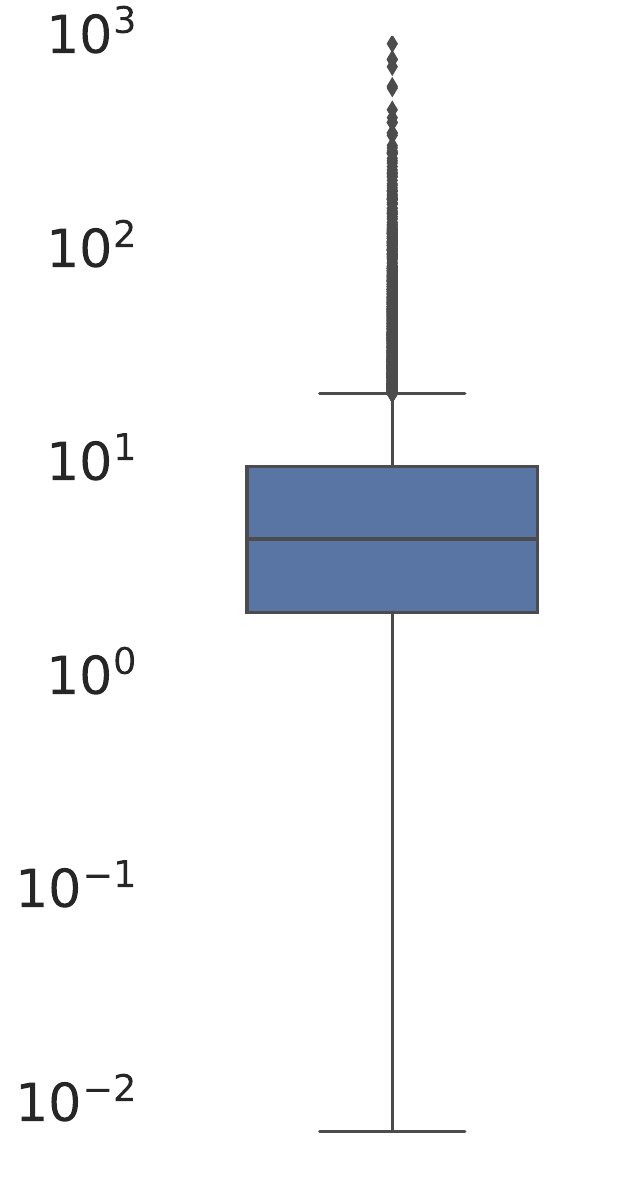} & \includegraphics[width=.2\textwidth]{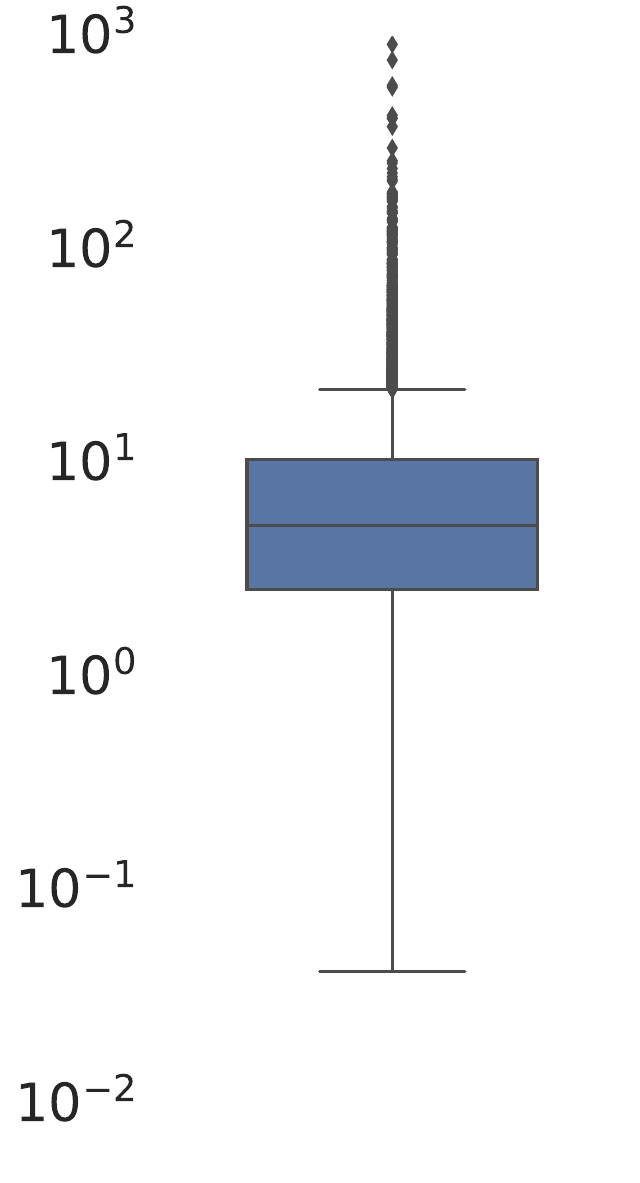} & \includegraphics[width=.2\textwidth]{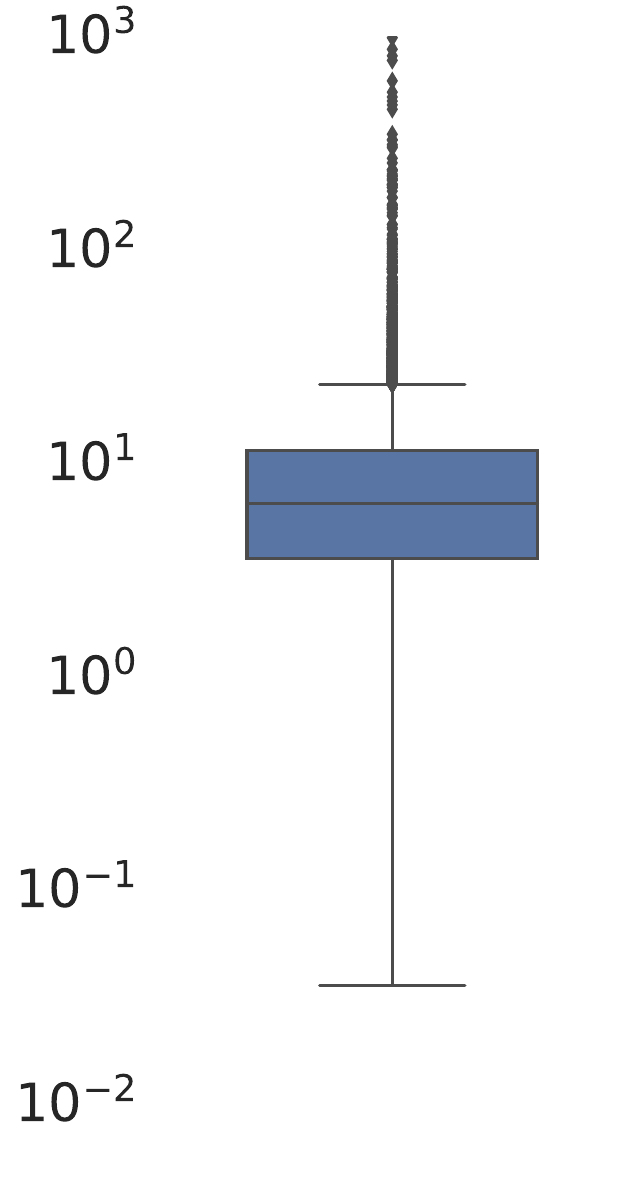} \\
\bottomrule
\end{tabular}
}
\caption{Intrinsic evaluation results for English and Arabic. Plots for other languages are in App.~\ref{app:score-ratios}.}
\label{fig:score-ratio-boxplots}
\end{figure}

\paragraph{Is \ourdata data harder to predict than original data?} 
To test Hypothesis 1, we check the distribution of score ratios $r_f$ for all scoring functions $f$. 
The results for mBERT and mGPT are shown in Fig.~\ref{fig:score-ratio-boxplots} and show that for all settings, the first quartile is larger than 1, i.e. the score of \ourdata sentences is higher than that of original sentences for the vast majority of the data.
This means that \ourdata data receives lower likelihood than original data and is therefore harder to predict, as expected.

\paragraph{Are scoring functions affected differently by nonce data?}
To answer this question, we compare the distributions of score ratios for all scoring functions, as displayed in Fig.~\ref{fig:score-ratio-boxplots}.
This question is answered in two parts, and both parts hold irrespective of whether outliers with $r_f>250$ are taken into account or not.
First, we test Hypothesis 2 by comparing the distributions of score ratios for ALMs and MLMs. 
Here, the picture is clear: $r_{PPL}$ are significantly higher than both $r_{PPPL_{l2r}}$ and $r_{PPPL}$, measured with a Wilcoxon signed-rank test. 
This result, and all following results, are significant at $p \ll 0.001$.
Second, we test Hypothesis 3 by comparing the MLM scoring functions $r_{PPPL_{l2r}}$ and $r_{PPPL}$. 
Here, the picture is less clear: 
All ratios are significantly different from each other according to the same test, 
but $f_{PPPL_{l2r}}$ has slightly lower quartiles than $r_{PPPL}$ for Arabic (in all other languages, the opposite is true). 
This means that for all languages except Arabic, $PPPL$ is less affected by \ourdata data than $PPPL_{l2r}$.

\paragraph{Differences across languages and models} 
The same trends were tested for monolingual Arabic and English models. The results are presented in App.~\ref{app:score-ratios}. 
In all monolingual settings, \ourdata data is harder to predict than original data (all first quartiles of ratio distributions are larger than 1). 
For differences between ratio predictions, the picture is more mixed: 
Regarding the difference between ALMs and MLMs, Arabic AraGPT2 \cite{antoun-etal-2021-aragpt2} shows higher ratio quartiles for $PPL$ than the MLM scoring functions of AraBERT \cite{antoun-etal-2020-arabert}.
For English, the MLM scoring functions of RoBERTa \cite{liu-etal-2019-roberta} are significantly higher than $PPL$ of GPT-2 \cite{radford-etal-2019-gpt2}. %
Regarding the difference between $PPPL$ and $PPPL_{l2r}$, the picture is the same as for multilingual models: 
For English, $PPPL_{l2r}$ is significantly higher than $PPPL$, and for Arabic, the opposite is true.

\subsection{Interim Summary and Discussion}

There are three main takeaways from this experiment. 
First, we find (as expected) that \ourdata is harder to predict than original data in all settings. 
Second, 
the ALM perplexity and two MLM scoring functions respond differently to \ourdata data: 
mBERT is affected less than mGPT, and within mBERT, $PPPL_{l2r}$ is affected more than $PPPL$ for all languages except Arabic (because of the availability of subword context). 
For monolingual models, we see a more mixed picture where for English, ALMs are affected less by \ourdata data than MLMs, and for Arabic, the trend that $PPPL_{l2r}$ is affected more than $PPPL$ is consistent.
These 
results suggest on one hand a special role of the Arabic language, and on the other hand point to the possibility that these differences are not systematic properties of model architectures and scoring functions, but rather depend on the training data and tokenization. 
However, as a general trend, we can hypothesize from the comparison between $PPPL$ and $PPPL_{l2r}$ that the inflated scores for multi-token words that \citet{kauf-ivanova-2023-better} found for $PPPL$ are carried over to \ourdata data: If $PPPL$ increases less than $PPPL_{l2r}$ for nonce data, this means that the increase in $PPPL$ is cushioned by the high predictability of multi-token words in $PPPL$. 
We measure the lexical diversity of \ourdata in App.~\ref{app:ttr}. 
For all languages, Type-Token-Ratio (TTR) decreases. 
This means that all our findings hold even though \ourdata is less lexically diverse than natural data.

\section{Nonce Dependency Probing}\label{sec:nonce-probing}

\todo{Make hypotheses and motivation explicit}
In this section, we measure the efficacy of \ourdata in decoding the syntactic knowledge learned in LM representations.
Concretely, we train DepProbe structural probes \cite{muller-eberstein-etal-2022-probing} and compare the performance on the standard (original) UD test sets with performance on \ourdata test sets. 
We choose DepProbe because none of its alternatives produces both directed and labeled dependency trees. %
We investigate to which extent DepProbe performance is influenced by lexical semantics rather than its desired task of predicting purely syntactic properties. 
With the assumption from previous work (Sec.~\ref{sec:related-work}) that a certain drop in performance on nonce test data is expected, we answer the following research questions:\\
\noindent RQ1: \textit{Is MLM and ALM performance affected differently when tested on \ourdata data?}\\
\noindent RQ2: \textit{Are the predictions of dependencies between tokens (edges) and dependency relations (edge labels) affected differently by \ourdata data?}\\
We answer these questions on all five languages, and present additional experiments to improve the robustness of our findings.

\subsection{Probing Model}

\citet{muller-eberstein-etal-2022-probing} presented DepProbe, a lightweight decoder for directed and labeled dependency trees. 
The model consists of two learnt linear transformations of the LM representations. 
Both transform $d_h$-dimensional word representations into vectors that highlight a feature of the dependency tree. 
$L$ %
is a sequence labeling classifier that predicts for each word the label of the incoming dependency edge. 
$B \in \mathbb{R}^{d_h \times b}$ projects the LM representation in a \textit{syntactic subspace} with $b < d_h$, where word vector distances mimic the distance between words in the tree. 
The dependency tree is constructed in a top-down fashion from both components, selecting the word as root to which $L$ assigns the highest probability of being the root. %
Details are provided in App.~\ref{app:depprobe-arch}. 

\subsection{Experimental setup}

\paragraph{Metrics} 
We report the same metrics as \citet{muller-eberstein-etal-2022-probing}, all of which are commonly used in the dependency parsing and probing literature. 
\relacc is the percentage of tokens for which the relation label of the incoming edge %
 is correctly predicted, %
\uas (unlabeled attachment score) is the percentage of tokens for which the head token is correctly predicted, and %
\las %
 (labeled attachment score) is the conjunction of \relacc and \uasNoSpace.

\paragraph{Hyperparameters}
All probe hyperparameters %
were 
taken from~\citet{muller-eberstein-etal-2022-probing} in order to replicate their results. 
For all models, the dimensionality of the syntactic subspace is $b=128$. 
We take the hidden representations of layers 6 (7) as input to DepProbe's distance (relation) component for all twelve-layered models. 
\citet{muller-eberstein-etal-2022-probing} trained probes on English data for all layers of mBERT and chose the best-performing layer for each component. 
We repeated this experiment to determine the best-performing layer for the 24-layered mGPT (App.~\ref{app:layer-dynamics}), and found that layer 12 performs best for both components. 

\paragraph{Baselines}
We 
use
three baselines to ensure that the syntactic information we probe for is 
(i) contextual in nature rather than dependent on word types, and 
(ii) learnt during LM pretraining rather than by the probe. 
To estimate how much information about the probing task is contained in (context-insensitive) token embeddings, we train DepProbe on mBERT's embedding layer. 
To estimate how much information about the probing task is acquired during pretraining of a LM, we probe the internal representations of two MLMs that share the architecture of mBERT but randomize either all parameters, or all parameters except the embedding matrix \cite{belinkov-2022-probing}. 
All trained probes outperform the baselines by large margins (minimal difference of 15.9 \las for mGPT on \ourdata data). 
All baseline results are presented in App.~\ref{app:baseline-scores}. 

\subsection{Results}
Tables~\ref{tab:depprobe-mlm-results}, \ref{tab:depprobe-alm-results} present the results.
$\Delta$ shows the performance drop between original and \ourdata test sets. 
In general, performance depends largely on treebank size. This trend is also visible in \citet{muller-eberstein-etal-2022-probing}, where Kendall's $\tau$ between \las and treebank size in 13 languages is 0.50 (p = 0.017). 

\begin{table}
    \centering
    \footnotesize

\begin{tabular}{l|rr|rr|rr}
\toprule
{} & \multicolumn{2}{c|}{\relacc} &  \multicolumn{2}{c|}{\uas} & \multicolumn{2}{c}{\las} \\
& orig & $\Delta$ & orig & $\Delta$ & orig & $\Delta$ \\
\midrule
ar &  82.8 &  5.5 &  63.1 &  7.3 &  55.5 &  8.3 \\
de &  92.7 &  1.7 &  84.4 &  3.4 &  80.2 &  4.0 \\
en &  87.1 &  2.3 &  74.2 &  3.8 &  67.9 &  4.1 \\
fr &  89.3 &  2.0 &  76.0 &  4.4 &  70.4 &  4.3 \\
ru &  88.2 &  1.5 &  75.4 &  3.2 &  69.2 &  3.1 \\
\bottomrule
\end{tabular}
    \caption{DepProbe results for mBERT. $\Delta$ shows the performance drop when nonce data is used.}
    \label{tab:depprobe-mlm-results}
\end{table}

\begin{table}
\centering\footnotesize

\begin{tabular}{l|rr|rr|rr}
\toprule
{} & \multicolumn{2}{c|}{\relacc} &  \multicolumn{2}{c|}{\uas} & \multicolumn{2}{c}{\las} \\
& orig & $\Delta$ & orig & $\Delta$ & orig & $\Delta$ \\
\midrule
ar &  85.2 &  7.7 &  59.5 &  12.6 &  53.4 &  13.5 \\
de &  92.6 &  3.6 &  84.2 &  6.8 &  79.9 &  8.1 \\
en &  82.8 &  5.2 &  58.9 &  5.5 &  52.2 &  6.5 \\
fr &  87.5 &  2.8 &  69.6 &  6.2 &  64.0 &  6.7 \\
ru &  88.7 &  3.0 &  75.6 &  5.4 &  69.6 &  6.0 \\
\bottomrule
\end{tabular}
    \caption{DepProbe results for mGPT. $\Delta$ shows the performance drop when nonce data is used.}
    \label{tab:depprobe-alm-results}
\end{table}

  \begin{table}
    \centering
    \footnotesize
    \begin{tabular}{l|rrrrrr}
\toprule
{} & \multicolumn{2}{c}{\relacc} &  \multicolumn{2}{c}{\uas} & \multicolumn{2}{c}{\las} \\
& orig & $\Delta$ & orig & $\Delta$ & orig & $\Delta$ \\
\midrule
roberta-base      & 85.9 & 2.5 & 70.9 & 3.6 & 64.0 & 4.0 \\
gpt2              & 82.4 & 4.5 & 53.1 & 3.6 & 47.1 & 4.6 \\
\bottomrule
\end{tabular}
    \caption{DepProbe monolingual results for English}   
    \label{tab:depprobe-monoling-results}
\end{table}

\paragraph{mBERT vs. mGPT (RQ1)} 
For all languages and settings except 
Arabic \relacc (both test sets), 
and Russian (all metrics, original data), 
mBERT outperforms mGPT in terms of absolute performance. 
In addition, $\Delta$ on \ourdata is larger for mGPT than for mBERT in almost all settings. 
The difference in model architectures is especially relevant for predicting the syntactic structure of nonce data, where context is a more important cue than for original data. 
The only exception to this is Russian, where mGPT outperforms mBERT in absolute performance on original data but not on \ourdataNoSpace. 
This is likely due to the distribution of languages in the pretraining data: mGPT's training data put an emphasis on Russian and regionally related languages \cite{shliazhko-etal-2022-mgpt}. 
The relatively low performance of both models on English, however, is most likely explained by the relatively small size of the English treebank.

\paragraph{Relation labeling vs. attachment (RQ2)}
DepProbe performance is measured in terms of the two components of the model: 
relation labeling (\relaccNoSpace) and attachment (\uasNoSpace). 
While the absolute performance varies, for both models and all languages, $\Delta$ is larger for the attachment component than for the relation labeling component. 
mGPT shows a larger $\Delta$ than mBERT for both components when directly comparing the absolute performance drop per language. 
In the most extreme case, the $\Delta$ in \uas for Arabic is 7.3 points for mBERT and 12.6 points for mGPT.
This points in the same direction as the performance of the embedding layer baseline: Relations are more easily predictable from lexical cues. %
Attachment, on the other hand, requires more contextual information, especially when the head is in the future context of the dependent. 
This finding is particularly insightful because it shows that probing for labeled and directed dependency trees highlights differences between ALMs and MLMs that are not captured when ignoring relations and directionality \cite[e.g.][]{eisape-etal-2022-probing}.

\paragraph{Monolingual results (RQ1,RQ2)}
We test the above hypotheses on two English 12-layered models: GPT-2 \cite{radford-etal-2019-gpt2} and RoBERTa \cite{liu-etal-2019-roberta}. %
The trends (Tab.~\ref{tab:depprobe-monoling-results}) are similar to the multilingual models: %
 RoBERTa outperforms GPT-2 on all metrics, and the performance drop is larger for GPT-2 than for RoBERTa (except for \uasNoSpace, where it %
  is the same). 
The only difference is that for GPT-2, the separate probe components show a behavior inverse to the multilingual models: 
\relacc shows a larger performance drop on \ourdata data (4.5) than \uas (3.6). However, this can be explained by the absolute performances of the GPT-2 probe, which are much higher for \relacc than for \uasNoSpace. 
This leaves less room for performance detriments on \ourdata data at attachment than at %
 relation labeling. %

\subsection{Additional experiments}

To test the robustness of our results, we conduct several additional experiments. 
First, we test the effect of different random seeds for the probe's initial weights and the selection of nonce words (App.~\ref{app:exp-random-seeds}). 
We find that performance is stable across random seeds. 
Concretely, the largest observed standard deviation across random seeds was 0.35 points \uas for English \ourdata data, and most of the other settings show much lower standard deviations. 
Second, we test the effect of ignoring dependency relations in nonce data creation, i.e. selecting replacements based only on POS tags and language-specific rules (App.~\ref{app:exp-nodeps}). 
The results show that this kind of nonce data produces worse probing results than \ourdata data, with some cross-lingual variation. 
Finally, we train DepProbe on all layers of mBERT and mGPT and compare the performance on English original and \ourdata data (App.~\ref{app:layer-dynamics}). 
Most importantly, this confirms that all other experiments were run on the best-performing layers. 
Additionally, this provides insights into the layerwise dynamics of both models.%

\subsection{Interim Summary}

The probing experiments show an overall performance drop on \ourdata data for both MLMs and ALMs. 
In general, probes for MLMs perform better than probes for ALMs on both original and \ourdataNoSpace , and the performance drop on \ourdata is higher for ALMs than for MLMs. 
This is likely due to the fact that ALM probe predictions are based on a representation with one-sided contextual information.\footnote{In a small-scale qualitative analysis, we find that this can explain many errors made by ALMs but not MLMs.}
All models show a larger peformance drop for attachment than for relation labeling, and the embedding layer baseline shows a relatively high performance on relation labeling. 
This points in the direction that relation labeling information is more easily predictable from lexical cues than attachment information and that, on the contrary, MLMs are better at predicting attachment information for nonce data than ALMs.  
Finally, when viewed in isolation, many \ourdata test sets show a high absolute performance, suggesting that probes indeed base their predictions on syntactic information.

\section{Discussion and Conclusion}\label{sec:conclusions}
\paragraph{Dataset} We presented \textit{Semantically Perturbed Universal Dependencies} (\ourdataNoSpace ), a framework for creating parallel nonsensical UD treebanks, 
and apply it to generate grammatical but nonsensical data in five languages.   
We present two use cases for \ourdataNoSpace , which contribute to the %
literature that relates LMs to the notions of grammaticality, acceptability and probability \cite{gulordava-etal-2018-colorless,sprouse-etal-2018-colorless,lau-etal-2017-grammaticality,lau-etal-2020-furiously}. 
Beyond that, \ourdata has several possible applications 
such as data augmentation for low-resource parsing or syntactic similarity benchmarks.

\paragraph{Scoring \ourdata with ALMs and MLMs}
In analyzing how token-level predictions are affected by \ourdataNoSpace,
we find that \ourdata data consistently shows higher scores (i.e., lower likelihood) than %
 original data on all models. 
This shows that, as desired, the %
lexical co-occurrence patterns of content words in \ourdata are perturbed. 
Of all scoring functions, the increase in ALM perplexity is highest. 
With our experiments, we contribute to the open question of how best to compute scores from MLMs by comparing $PPPL$ with subword-aware $PPPL_{l2r}$ \cite{kauf-ivanova-2023-better}. 
While we are mostly able to support \cite{kauf-ivanova-2023-better}'s argument that $PPPL_{l2r}$ is empirically more similar to $PPL$ than $PPPL$ is, we find that the response of ALMs and MLMs to \ourdata data is significantly different. 
This indicates a more fundamental difference between how ALMs and MLMs score sentences: Bidirectional context in MLMs makes it easier to identify the syntactic properties of a predicted word, even in nonce sentences. 

\paragraph{Probing ALMs and MLMs for \ourdata{} trees} 
We probe different models for labeled and directed dependency trees on \ourdata using the DepProbe framework \cite{muller-eberstein-etal-2022-probing}. 
We find that probe performance drops (compared to original test sets) for all languages and both model architectures when using \ourdata test sets. However, a majority of the probe performance is maintained. 
The ALM mGPT has a lower 
overall 
performance than mBERT, as well as a higher performance drop on \ourdata data. 
For both 
architectures, attachment contributes more to the performance drop than relation labeling does. %
These findings confirm the results in \citet{hall-maudslay-cotterell-2021-syntactic} and \citet{arps-etal-2022-probing}. 
On the contrary, \citet{eisape-etal-2022-probing} found that GPT-2 performs on par with MLMs on structural probing of unlabeled %
 undirected dependency trees. 
This contrast highlights the importance of probe architecture and task design.

\paragraph{Linguistic Information and LMs} \citet{sinha-etal-2021-masked} found that MLMs rely on higher-order co-occurrence statistics for many tasks, and can maintain high performance without word order information. We view the same problem from a different angle: Our scoring results show that, expectedly, LM predictions are optimized for co-occurrence information. 
Probing performance, however, can be maintained to a high degree even if co-occurrence at the lexical level is disrupted. 
Our comparison of MLMs and ALMs suggests that both architectures process 
nonce data differently, and MLMs are better at identifying syntactic structure in \ourdata data than ALMs.

\section*{Limitations}

\paragraph{Model instances and architectures}
All our direct comparisons between model architectures are limited by the fact that we compare models with different hyperparameters, training data, and tokenizers. 
To quantify the effect of each of these components on the syntactic learning requires availability of models trained using various combination of these settings. 

\paragraph{Data}
While we put significant effort into assuring the quality of the generated data, we did not conduct a large-scale human evaluation, e.g. via crowd-sourced grammaticality judgments. 
This means that while we are confident that grammatical errors in \ourdata are rare, rare errors might exist that have potentially severe effects on individual predictions.

\paragraph{Formulations of scoring functions}%
The literature varies in the way in which scoring functions are applied. 
Assume that $\sigma = \sum_{i=0}^n \log p(x_i|context)$ is the sum of log likelihoods in a scoring function such as $PPL$.
We follow \citet{salazar-etal-2020-masked} and divide $\sigma$ by sequence length and exponentiate ($exp(-(1/n) \sigma)$), as defined in Eq.~\ref{eq:ppl}. 
\citet{kauf-ivanova-2023-better}, on the contrary, base their experiments directly on $\sigma$. 
Due to the nonlinear differences between both versions, it is possible that not all of our findings on scoring functions are transferable to raw log likelihood scores. 
We do not expect this to affect our main findings, especially since our comparison of the exponentiated versions of $PPPL$ and $PPPL_{l2r}$ shows trends that are consistent with \citet{kauf-ivanova-2023-better}'s findings on English data, the language they focus on. 
However, we deem a more detailed comparison of scoring functions with and without exponentiation is necessary to draw more general conclusions, even though this is beyond the scope of this paper. 

\paragraph{Structural Probing and Properties of Explanations} 
In the following, we discuss limitations of our probing experiments in terms of the properties of explanations defined by \citet{lyu-etal-2023-faithful}. 
Several of their criteria are met: 
First, our experiments are \textit{plausible} (the resulting dependency trees are intuitive to humans), 
Second, the critierion of \textit{Input sensitivity} relates closely to the interaction of syntax and semantics in probing which we investigate. 
The definition is that ``An explanation should be sensitive (resp. insensitive) to changes in the input that influence (resp. do not influence) the prediction'' \cite[p.~6]{lyu-etal-2023-faithful}.
This criterion is met because (i) we directly investigate the effect of changing the semantic content of the probe and model input, and (ii) because DepProbe is generally sensitive to changes in syntax. 
Third, the criterion of \textit{Faithfulness} states that ``an explanation should accurately represent the reasoning process behind the model's prediction'' \cite[p.~1]{lyu-etal-2023-faithful}. 
Since we do not investigate model predictions but probe predictions, this criterion is not directly applicable. 
However, we can say that the probe predictions need not be faithful, because DepProbe learns to predict syntactic structure in a supervised fashion, and does not necessarily decode the syntactic generalizations that the model has learnt. 
Finally, \textit{Completeness} and \textit{Minimality} do not apply because they refer to explaining contributing factors for model predictions rather than probe predictions.

\bibliography{anthology,custom}

\begin{thebibliography}{64}
\expandafter\ifx\csname natexlab\endcsname\relax\def\natexlab#1{#1}\fi

\bibitem[{Antoun et~al.(2020)Antoun, Baly, and Hajj}]{antoun-etal-2020-arabert}
Wissam Antoun, Fady Baly, and Hazem Hajj. 2020.
\newblock \href {https://aclanthology.org/2020.osact-1.2} {{A}ra{BERT}:
  Transformer-based model for {A}rabic language understanding}.
\newblock In \emph{Proceedings of the 4th Workshop on Open-Source Arabic
  Corpora and Processing Tools, with a Shared Task on Offensive Language
  Detection}, pages 9--15, Marseille, France. European Language Resource
  Association.

\bibitem[{Antoun et~al.(2021)Antoun, Baly, and Hajj}]{antoun-etal-2021-aragpt2}
Wissam Antoun, Fady Baly, and Hazem Hajj. 2021.
\newblock \href {https://aclanthology.org/2021.wanlp-1.21} {{A}ra{GPT}2:
  Pre-trained transformer for {A}rabic language generation}.
\newblock In \emph{Proceedings of the Sixth Arabic Natural Language Processing
  Workshop}, pages 196--207, Kyiv, Ukraine (Virtual). Association for
  Computational Linguistics.

\bibitem[{Arps et~al.(2022)Arps, Samih, Kallmeyer, and
  Sajjad}]{arps-etal-2022-probing}
David Arps, Younes Samih, Laura Kallmeyer, and Hassan Sajjad. 2022.
\newblock \href {https://doi.org/10.18653/v1/2022.findings-emnlp.502} {Probing
  for constituency structure in neural language models}.
\newblock In \emph{Findings of the Association for Computational Linguistics:
  EMNLP 2022}, pages 6738--6757, Abu Dhabi, United Arab Emirates. Association
  for Computational Linguistics.

\bibitem[{Belinkov(2022)}]{belinkov-2022-probing}
Yonatan Belinkov. 2022.
\newblock \href {https://doi.org/10.1162/coli_a_00422} {Probing classifiers:
  Promises, shortcomings, and advances}.
\newblock \emph{Computational Linguistics}, 48(1):207--219.

\bibitem[{Belinkov et~al.(2017)Belinkov, Durrani, Dalvi, Sajjad, and
  Glass}]{belinkov:2017:acl}
Yonatan Belinkov, Nadir Durrani, Fahim Dalvi, Hassan Sajjad, and James Glass.
  2017.
\newblock \href
  {https://aclanthology.coli.uni-saarland.de/pdf/P/P17/P17-1080.pdf} {{What do
  Neural Machine Translation Models Learn about Morphology?}}
\newblock In \emph{Proceedings of the 55th Annual Meeting of the Association
  for Computational Linguistics (ACL)}, Vancouver. Association for
  Computational Linguistics.

\bibitem[{Blevins et~al.(2018)Blevins, Levy, and
  Zettlemoyer}]{blevins-etal-2018-deep}
Terra Blevins, Omer Levy, and Luke Zettlemoyer. 2018.
\newblock \href {https://doi.org/10.18653/v1/P18-2003} {Deep {RNN}s encode soft
  hierarchical syntax}.
\newblock In \emph{Proceedings of the 56th Annual Meeting of the Association
  for Computational Linguistics (Volume 2: Short Papers)}, pages 14--19,
  Melbourne, Australia. Association for Computational Linguistics.

\bibitem[{Borges~V{\"o}lker et~al.(2019)Borges~V{\"o}lker, Wendt, Hennig, and
  K{\"o}hn}]{borges-volker-etal-2019-hdt}
Emanuel Borges~V{\"o}lker, Maximilian Wendt, Felix Hennig, and Arne K{\"o}hn.
  2019.
\newblock \href {https://doi.org/10.18653/v1/W19-8006} {{HDT}-{UD}: A very
  large {U}niversal {D}ependencies treebank for {G}erman}.
\newblock In \emph{Proceedings of the Third Workshop on Universal Dependencies
  (UDW, SyntaxFest 2019)}, pages 46--57, Paris, France. Association for
  Computational Linguistics.

\bibitem[{Chang and Bergen(2023)}]{chang-bergen-2023-language}
Tyler~A. Chang and Benjamin~K. Bergen. 2023.
\newblock \href {https://doi.org/10.1162/coli_a_00492} {{Language Model
  Behavior: A Comprehensive Survey}}.
\newblock \emph{Computational Linguistics}.

\bibitem[{Chi et~al.(2020)Chi, Hewitt, and Manning}]{chi-etal-2020-finding}
Ethan~A. Chi, John Hewitt, and Christopher~D. Manning. 2020.
\newblock \href {https://doi.org/10.18653/v1/2020.acl-main.493} {Finding
  universal grammatical relations in multilingual {BERT}}.
\newblock In \emph{Proceedings of the 58th Annual Meeting of the Association
  for Computational Linguistics}, pages 5564--5577, Online. Association for
  Computational Linguistics.

\bibitem[{Chomsky(1957)}]{chomsky-1957-syntactic-structures}
Noam Chomsky. 1957.
\newblock \emph{{Syntactic Structures}}.
\newblock Mouton and Co., The Hague.

\bibitem[{Darwish and Mubarak(2016)}]{darwish-mubarak-2016-farasa}
Kareem Darwish and Hamdy Mubarak. 2016.
\newblock \href {https://aclanthology.org/L16-1170} {{F}arasa: A new fast and
  accurate {A}rabic word segmenter}.
\newblock In \emph{Proceedings of the Tenth International Conference on
  Language Resources and Evaluation ({LREC}'16)}, pages 1070--1074,
  Portoro{\v{z}}, Slovenia. European Language Resources Association (ELRA).

\bibitem[{de~Marneffe et~al.(2021)de~Marneffe, Manning, Nivre, and
  Zeman}]{de-marneffe-etal-2021-universal}
Marie-Catherine de~Marneffe, Christopher~D. Manning, Joakim Nivre, and Daniel
  Zeman. 2021.
\newblock \href {https://doi.org/10.1162/coli_a_00402} {{U}niversal
  {D}ependencies}.
\newblock \emph{Computational Linguistics}, 47(2):255--308.

\bibitem[{Dehouck and
  G{\'o}mez-Rodr{\'\i}guez(2020)}]{dehouck-gomez-rodriguez-2020-data}
Mathieu Dehouck and Carlos G{\'o}mez-Rodr{\'\i}guez. 2020.
\newblock \href {https://doi.org/10.18653/v1/2020.coling-main.339} {Data
  augmentation via subtree swapping for dependency parsing of low-resource
  languages}.
\newblock In \emph{Proceedings of the 28th International Conference on
  Computational Linguistics}, pages 3818--3830, Barcelona, Spain (Online).
  International Committee on Computational Linguistics.

\bibitem[{Devlin et~al.(2019)Devlin, Chang, Lee, and
  Toutanova}]{devlin-etal-2019-bert}
Jacob Devlin, Ming-Wei Chang, Kenton Lee, and Kristina Toutanova. 2019.
\newblock \href {https://doi.org/10.18653/v1/N19-1423} {{BERT}: Pre-training of
  deep bidirectional transformers for language understanding}.
\newblock In \emph{Proceedings of the 2019 Conference of the North {A}merican
  Chapter of the Association for Computational Linguistics: Human Language
  Technologies, Volume 1 (Long and Short Papers)}, pages 4171--4186,
  Minneapolis, Minnesota. Association for Computational Linguistics.

\bibitem[{Droganova et~al.(2018)Droganova, Lyashevskaya, and
  Zeman}]{droganova-etal-2018-data}
Kira Droganova, Olga Lyashevskaya, and Daniel Zeman. 2018.
\newblock {Data Conversion and Consistency of Monolingual Corpora: Russian UD
  Treebanks}.
\newblock In \emph{Proceedings of the 17th international workshop on treebanks
  and linguistic theories (tlt 2018)}, 155, pages 53--66. Link{\"o}ping
  University Electronic Press Link{\"o}ping, Sweden.

\bibitem[{Dryer and Haspelmath(2013)}]{wals}
Matthew~S. Dryer and Martin Haspelmath, editors. 2013.
\newblock \href {https://wals.info/} {\emph{WALS Online}}.
\newblock Max Planck Institute for Evolutionary Anthropology, Leipzig.

\bibitem[{Durrani et~al.(2024)Durrani, Dalvi, and
  Sajjad}]{durrani_neuron_jmlr_2023}
Nadir Durrani, Fahim Dalvi, and Hassan Sajjad. 2024.
\newblock Discovering salient neurons in deep nlp models.
\newblock \emph{J. Mach. Learn. Res.}, 24(1).

\bibitem[{Eisape et~al.(2022)Eisape, Gangireddy, Levy, and
  Kim}]{eisape-etal-2022-probing}
Tiwalayo Eisape, Vineet Gangireddy, Roger Levy, and Yoon Kim. 2022.
\newblock \href {https://doi.org/10.18653/v1/2022.findings-emnlp.203} {Probing
  for incremental parse states in autoregressive language models}.
\newblock In \emph{Findings of the Association for Computational Linguistics:
  EMNLP 2022}, pages 2801--2813, Abu Dhabi, United Arab Emirates. Association
  for Computational Linguistics.

\bibitem[{Guillaume et~al.(2019)Guillaume, de~Marneffe, and
  Perrier}]{guillaume-etal-2019-conversion}
Bruno Guillaume, Marie-Catherine de~Marneffe, and Guy Perrier. 2019.
\newblock \href {https://aclanthology.org/2019.tal-2.4} {Conversion et
  am{\'e}liorations de corpus du fran{\c{c}}ais annot{\'e}s en {U}niversal
  {D}ependencies [conversion and improvement of {U}niversal {D}ependencies
  {F}rench corpora]}.
\newblock \emph{Traitement Automatique des Langues}, 60(2):71--95.

\bibitem[{Gulordava et~al.(2018)Gulordava, Bojanowski, Grave, Linzen, and
  Baroni}]{gulordava-etal-2018-colorless}
Kristina Gulordava, Piotr Bojanowski, Edouard Grave, Tal Linzen, and Marco
  Baroni. 2018.
\newblock \href {https://doi.org/10.18653/v1/N18-1108} {Colorless green
  recurrent networks dream hierarchically}.
\newblock In \emph{Proceedings of the 2018 Conference of the North {A}merican
  Chapter of the Association for Computational Linguistics: Human Language
  Technologies, Volume 1 (Long Papers)}, pages 1195--1205, New Orleans,
  Louisiana. Association for Computational Linguistics.

\bibitem[{Hajic et~al.(2009)Hajic, Smrz, Zemánek, Pajas, Šnaidauf, Beška,
  Kráčmar, and Hassanová}]{hajic-etal-2009-praguearabic}
Jan Hajic, Otakar Smrz, Petr Zemánek, Petr Pajas, Jan Šnaidauf, Emanuel
  Beška, Jakub Kráčmar, and Kamila Hassanová. 2009.
\newblock {Prague Arabic Dependency Treebank 1.0}.

\bibitem[{Hennigen et~al.(2020)Hennigen, Williams, and
  Cotterell}]{torroba-hennigen-etal-2020-intrinsic}
Lucas~Torroba Hennigen, Adina Williams, and Ryan Cotterell. 2020.
\newblock \href {https://doi.org/10.18653/v1/2020.emnlp-main.15} {Intrinsic
  probing through dimension selection}.
\newblock In \emph{Proceedings of the 2020 Conference on Empirical Methods in
  Natural Language Processing (EMNLP)}, pages 197--216, Online. Association for
  Computational Linguistics.

\bibitem[{Hewitt and Manning(2019)}]{hewitt-manning-2019-structural}
John Hewitt and Christopher~D. Manning. 2019.
\newblock \href {https://doi.org/10.18653/v1/N19-1419} {{A} structural probe
  for finding syntax in word representations}.
\newblock In \emph{Proceedings of the 2019 Conference of the North {A}merican
  Chapter of the Association for Computational Linguistics: Human Language
  Technologies, Volume 1 (Long and Short Papers)}, pages 4129--4138,
  Minneapolis, Minnesota. Association for Computational Linguistics.

\bibitem[{Hupkes et~al.(2018)Hupkes, Veldhoen, and
  Zuidema}]{hupkes-etal-2018-visualisation}
Dieuwke Hupkes, Sara Veldhoen, and Willem Zuidema. 2018.
\newblock Visualisation and 'diagnostic classifiers' reveal how recurrent and
  recursive neural networks process hierarchical structure.
\newblock \emph{Journal of Artificial Intelligence Research}, 61:907--926.

\bibitem[{Kauf and Ivanova(2023)}]{kauf-ivanova-2023-better}
Carina Kauf and Anna Ivanova. 2023.
\newblock \href {https://doi.org/10.18653/v1/2023.acl-short.80} {A better way
  to do masked language model scoring}.
\newblock In \emph{Proceedings of the 61st Annual Meeting of the Association
  for Computational Linguistics (Volume 2: Short Papers)}, pages 925--935,
  Toronto, Canada. Association for Computational Linguistics.

\bibitem[{Kauf et~al.(2023)Kauf, Tuckute, Levy, Andreas, and
  Fedorenko}]{kauf-etal-2023-lexical}
Carina Kauf, Greta Tuckute, Roger Levy, Jacob Andreas, and Evelina Fedorenko.
  2023.
\newblock \href {https://doi.org/10.1162/nol_a_00116} {{Lexical-Semantic
  Content, Not Syntactic Structure, Is the Main Contributor to ANN-Brain
  Similarity of fMRI Responses in the Language Network}}.
\newblock \emph{Neurobiology of Language}.

\bibitem[{K{\"o}hler(2003)}]{koehler-2003-typetoken}
Reinhard K{\"o}hler. 2003.
\newblock \href {https://doi.org/10.1007/978-3-322-81289-6_8} {\emph{{Zur
  Type-Token-Ratio syntaktischer Einheiten: Eine
  quantitativ-korpuslinguistische Studie}}}, pages 93--101. Deutscher
  Universit{\"a}tsverlag, Wiesbaden.

\bibitem[{Kulmizev and Nivre(2022)}]{kulmizev-nivre-2022-schrodingers}
Artur Kulmizev and Joakim Nivre. 2022.
\newblock \href {https://doi.org/10.3389/frai.2022.796788} {{Schrödinger's
  tree—On syntax and neural language models}}.
\newblock \emph{Frontiers in Artificial Intelligence}, 5.

\bibitem[{Kulmizev et~al.(2020)Kulmizev, Ravishankar, Abdou, and
  Nivre}]{kulmizev-etal-2020-neural}
Artur Kulmizev, Vinit Ravishankar, Mostafa Abdou, and Joakim Nivre. 2020.
\newblock \href {https://doi.org/10.18653/v1/2020.acl-main.375} {Do neural
  language models show preferences for syntactic formalisms?}
\newblock In \emph{Proceedings of the 58th Annual Meeting of the Association
  for Computational Linguistics}, pages 4077--4091, Online. Association for
  Computational Linguistics.

\bibitem[{Kunz and Kuhlmann(2020)}]{kunz-kuhlmann-2020-classifier}
Jenny Kunz and Marco Kuhlmann. 2020.
\newblock \href {https://doi.org/10.18653/v1/2020.coling-main.450} {Classifier
  probes may just learn from linear context features}.
\newblock In \emph{Proceedings of the 28th International Conference on
  Computational Linguistics}, pages 5136--5146, Barcelona, Spain (Online).
  International Committee on Computational Linguistics.

\bibitem[{Kunz and Kuhlmann(2021)}]{kunz-kuhlmann-2021-test}
Jenny Kunz and Marco Kuhlmann. 2021.
\newblock \href {https://doi.org/10.18653/v1/2021.blackboxnlp-1.2} {Test harder
  than you train: Probing with extrapolation splits}.
\newblock In \emph{Proceedings of the Fourth BlackboxNLP Workshop on Analyzing
  and Interpreting Neural Networks for NLP}, pages 15--25, Punta Cana,
  Dominican Republic. Association for Computational Linguistics.

\bibitem[{Lasri et~al.(2022{\natexlab{a}})Lasri, Lenci, and
  Poibeau}]{lasri-etal-2022-bert}
Karim Lasri, Alessandro Lenci, and Thierry Poibeau. 2022{\natexlab{a}}.
\newblock \href {https://doi.org/10.18653/v1/2022.findings-acl.181} {Does
  {BERT} really agree? fine-grained analysis of lexical dependence on a
  syntactic task}.
\newblock In \emph{Findings of the Association for Computational Linguistics:
  ACL 2022}, pages 2309--2315, Dublin, Ireland. Association for Computational
  Linguistics.

\bibitem[{Lasri et~al.(2022{\natexlab{b}})Lasri, Seminck, Lenci, and
  Poibeau}]{lasri-etal-2022-subject}
Karim Lasri, Olga Seminck, Alessandro Lenci, and Thierry Poibeau.
  2022{\natexlab{b}}.
\newblock \href {https://aclanthology.org/2022.coling-1.4} {Subject verb
  agreement error patterns in meaningless sentences: Humans vs. {BERT}}.
\newblock In \emph{Proceedings of the 29th International Conference on
  Computational Linguistics}, pages 37--43, Gyeongju, Republic of Korea.
  International Committee on Computational Linguistics.

\bibitem[{Lau et~al.(2020)Lau, Armendariz, Lappin, Purver, and
  Shu}]{lau-etal-2020-furiously}
Jey~Han Lau, Carlos Armendariz, Shalom Lappin, Matthew Purver, and Chang Shu.
  2020.
\newblock \href {https://doi.org/10.1162/tacl_a_00315} {How furiously can
  colorless green ideas sleep? sentence acceptability in context}.
\newblock \emph{Transactions of the Association for Computational Linguistics},
  8:296--310.

\bibitem[{Lau et~al.(2017)Lau, Clark, and
  Lappin}]{lau-etal-2017-grammaticality}
Jey~Han Lau, Alexander Clark, and Shalom Lappin. 2017.
\newblock \href {https://doi.org/https://doi.org/10.1111/cogs.12414}
  {{Grammaticality, Acceptability, and Probability: A Probabilistic View of
  Linguistic Knowledge}}.
\newblock \emph{Cognitive Science}, 41(5):1202--1241.

\bibitem[{Linzen and Baroni(2021)}]{linzen-baroni-2021-syntactic}
Tal Linzen and Marco Baroni. 2021.
\newblock \href {https://doi.org/10.1146/annurev-linguistics-032020-051035}
  {Syntactic structure from deep learning}.
\newblock \emph{Annual Review of Linguistics}, 7(1):195--212.

\bibitem[{Liu et~al.(2019)Liu, Ott, Goyal, Du, Joshi, Chen, Levy, Lewis,
  Zettlemoyer, and Stoyanov}]{liu-etal-2019-roberta}
Yinhan Liu, Myle Ott, Naman Goyal, Jingfei Du, Mandar Joshi, Danqi Chen, Omer
  Levy, Mike Lewis, Luke Zettlemoyer, and Veselin Stoyanov. 2019.
\newblock \href {http://arxiv.org/abs/1907.11692} {{RoBERTa: A Robustly
  Optimized BERT Pretraining Approach}}.

\bibitem[{Lyu et~al.(2024)Lyu, Apidianaki, and
  Callison-Burch}]{lyu-etal-2023-faithful}
Qing Lyu, Marianna Apidianaki, and Chris Callison-Burch. 2024.
\newblock \href {https://doi.org/10.1162/coli_a_00511} {{Towards Faithful Model
  Explanation in NLP: A Survey}}.
\newblock \emph{Computational Linguistics}.

\bibitem[{Mahowald et~al.(2023)Mahowald, Ivanova, Blank, Kanwisher, Tenenbaum,
  and Fedorenko}]{mahowald-etal-2023-dissociating}
Kyle Mahowald, Anna~A. Ivanova, Idan~Asher Blank, Nancy Kanwisher, Joshua~B.
  Tenenbaum, and Evelina Fedorenko. 2023.
\newblock \href {https://doi.org/10.48550/ARXIV.2301.06627} {Dissociating
  language and thought in large language models: a cognitive perspective}.
\newblock \emph{CoRR}, abs/2301.06627.

\bibitem[{Manning et~al.(2020)Manning, Clark, Hewitt, Khandelwal, and
  Levy}]{manning-etal-2020-emergent}
Christopher~D. Manning, Kevin Clark, John Hewitt, Urvashi Khandelwal, and Omer
  Levy. 2020.
\newblock \href {https://doi.org/10.1073/pnas.1907367117} {Emergent linguistic
  structure in artificial neural networks trained by self-supervision}.
\newblock \emph{Proceedings of the National Academy of Sciences},
  117(48):30046--30054.

\bibitem[{Marcus et~al.(1993)Marcus, Santorini, and
  Marcinkiewicz}]{marcus-etal-1993-building}
Mitchell~P. Marcus, Beatrice Santorini, and Mary~Ann Marcinkiewicz. 1993.
\newblock \href {https://aclanthology.org/J93-2004} {Building a large annotated
  corpus of {E}nglish: The {P}enn {T}reebank}.
\newblock \emph{Computational Linguistics}, 19(2):313--330.

\bibitem[{Marvin and Linzen(2018)}]{marvin-linzen-2018-targeted}
Rebecca Marvin and Tal Linzen. 2018.
\newblock \href {https://doi.org/10.18653/v1/D18-1151} {Targeted syntactic
  evaluation of language models}.
\newblock In \emph{Proceedings of the 2018 Conference on Empirical Methods in
  Natural Language Processing}, pages 1192--1202, Brussels, Belgium.
  Association for Computational Linguistics.

\bibitem[{Maudslay and
  Cotterell(2021)}]{hall-maudslay-cotterell-2021-syntactic}
Rowan~Hall Maudslay and Ryan Cotterell. 2021.
\newblock \href {https://doi.org/10.18653/v1/2021.naacl-main.11} {Do syntactic
  probes probe syntax? experiments with jabberwocky probing}.
\newblock In \emph{Proceedings of the 2021 Conference of the North American
  Chapter of the Association for Computational Linguistics: Human Language
  Technologies}, pages 124--131, Online. Association for Computational
  Linguistics.

\bibitem[{Miaschi et~al.(2021)Miaschi, Brunato, Dell{'}Orletta, and
  Venturi}]{miaschi-etal-2021-makes}
Alessio Miaschi, Dominique Brunato, Felice Dell{'}Orletta, and Giulia Venturi.
  2021.
\newblock \href {https://doi.org/10.18653/v1/2021.deelio-1.5} {What makes my
  model perplexed? a linguistic investigation on neural language models
  perplexity}.
\newblock In \emph{Proceedings of Deep Learning Inside Out (DeeLIO): The 2nd
  Workshop on Knowledge Extraction and Integration for Deep Learning
  Architectures}, pages 40--47, Online. Association for Computational
  Linguistics.

\bibitem[{M{\"u}ller-Eberstein et~al.(2022)M{\"u}ller-Eberstein, van~der Goot,
  and Plank}]{muller-eberstein-etal-2022-probing}
Max M{\"u}ller-Eberstein, Rob van~der Goot, and Barbara Plank. 2022.
\newblock \href {https://doi.org/10.18653/v1/2022.acl-long.532} {Probing for
  labeled dependency trees}.
\newblock In \emph{Proceedings of the 60th Annual Meeting of the Association
  for Computational Linguistics (Volume 1: Long Papers)}, pages 7711--7726,
  Dublin, Ireland. Association for Computational Linguistics.

\bibitem[{Nagy et~al.(2023)Nagy, Lakatos, Barta, and
  {\'A}cs}]{nagy-etal-2023-data}
Attila Nagy, Dorina Lakatos, Botond Barta, and Judit {\'A}cs. 2023.
\newblock \href {https://aclanthology.org/2023.ranlp-1.82} {{T}ree{S}wap: Data
  augmentation for machine translation via dependency subtree swapping}.
\newblock In \emph{Proceedings of the 14th International Conference on Recent
  Advances in Natural Language Processing}, pages 759--768, Varna, Bulgaria.
  INCOMA Ltd., Shoumen, Bulgaria.

\bibitem[{Newman et~al.(2021)Newman, Ang, Gong, and
  Hewitt}]{newman-etal-2021-refining}
Benjamin Newman, Kai-Siang Ang, Julia Gong, and John Hewitt. 2021.
\newblock \href {https://doi.org/10.18653/v1/2021.naacl-main.290} {Refining
  targeted syntactic evaluation of language models}.
\newblock In \emph{Proceedings of the 2021 Conference of the North American
  Chapter of the Association for Computational Linguistics: Human Language
  Technologies}, pages 3710--3723, Online. Association for Computational
  Linguistics.

\bibitem[{Papadimitriou et~al.(2022)Papadimitriou, Futrell, and
  Mahowald}]{papadimitriou-etal-2022-classifying-grammatical}
Isabel Papadimitriou, Richard Futrell, and Kyle Mahowald. 2022.
\newblock \href {https://doi.org/10.18653/v1/2022.acl-short.71} {When
  classifying grammatical role, {BERT} doesn{'}t care about word order...
  except when it matters}.
\newblock In \emph{Proceedings of the 60th Annual Meeting of the Association
  for Computational Linguistics (Volume 2: Short Papers)}, pages 636--643,
  Dublin, Ireland. Association for Computational Linguistics.

\bibitem[{Radford et~al.(2019)Radford, Wu, Child, Luan, Amodei, Sutskever
  et~al.}]{radford-etal-2019-gpt2}
Alec Radford, Jeffrey Wu, Rewon Child, David Luan, Dario Amodei, Ilya
  Sutskever, et~al. 2019.
\newblock Language models are unsupervised multitask learners.
\newblock \emph{OpenAI blog}.

\bibitem[{Ravfogel et~al.(2020)Ravfogel, Elazar, Goldberger, and
  Goldberg}]{ravfogel-etal-2020-unsupervised}
Shauli Ravfogel, Yanai Elazar, Jacob Goldberger, and Yoav Goldberg. 2020.
\newblock \href {https://doi.org/10.18653/v1/2020.blackboxnlp-1.9}
  {Unsupervised distillation of syntactic information from contextualized word
  representations}.
\newblock In \emph{Proceedings of the Third BlackboxNLP Workshop on Analyzing
  and Interpreting Neural Networks for NLP}, pages 91--106, Online. Association
  for Computational Linguistics.

\bibitem[{Sagot(2018)}]{sagot-2018-multilingual}
Beno{\^\i}t Sagot. 2018.
\newblock \href {https://aclanthology.org/L18-1292} {A multilingual collection
  of {C}o{NLL}-{U}-compatible morphological lexicons}.
\newblock In \emph{Proceedings of the Eleventh International Conference on
  Language Resources and Evaluation ({LREC} 2018)}, Miyazaki, Japan. European
  Language Resources Association (ELRA).

\bibitem[{{\c{S}}ahin and Steedman(2018)}]{sahin-steedman-2018-data}
G{\"o}zde~G{\"u}l {\c{S}}ahin and Mark Steedman. 2018.
\newblock \href {https://doi.org/10.18653/v1/D18-1545} {Data augmentation via
  dependency tree morphing for low-resource languages}.
\newblock In \emph{Proceedings of the 2018 Conference on Empirical Methods in
  Natural Language Processing}, pages 5004--5009, Brussels, Belgium.
  Association for Computational Linguistics.

\bibitem[{Sajjad et~al.(2022)Sajjad, Durrani, and
  Dalvi}]{sajjad-etal-2022-neuron}
Hassan Sajjad, Nadir Durrani, and Fahim Dalvi. 2022.
\newblock \href {https://doi.org/10.1162/tacl_a_00519} {Neuron-level
  interpretation of deep {NLP} models: A survey}.
\newblock \emph{Transactions of the Association for Computational Linguistics},
  10:1285--1303.

\bibitem[{Salazar et~al.(2020)Salazar, Liang, Nguyen, and
  Kirchhoff}]{salazar-etal-2020-masked}
Julian Salazar, Davis Liang, Toan~Q. Nguyen, and Katrin Kirchhoff. 2020.
\newblock \href {https://doi.org/10.18653/v1/2020.acl-main.240} {Masked
  language model scoring}.
\newblock In \emph{Proceedings of the 58th Annual Meeting of the Association
  for Computational Linguistics}, pages 2699--2712, Online. Association for
  Computational Linguistics.

\bibitem[{Shliazhko et~al.(2024)Shliazhko, Fenogenova, Tikhonova, Kozlova,
  Mikhailov, and Shavrina}]{shliazhko-etal-2022-mgpt}
Oleh Shliazhko, Alena Fenogenova, Maria Tikhonova, Anastasia Kozlova, Vladislav
  Mikhailov, and Tatiana Shavrina. 2024.
\newblock \href {https://doi.org/10.1162/tacl_a_00633} {{mGPT: Few-Shot
  Learners Go Multilingual}}.
\newblock \emph{Transactions of the Association for Computational Linguistics},
  12:58--79.

\bibitem[{Silveira et~al.(2014)Silveira, Dozat, de~Marneffe, Bowman, Connor,
  Bauer, and Manning}]{silveira-etal-2014-gold}
Natalia Silveira, Timothy Dozat, Marie-Catherine de~Marneffe, Samuel Bowman,
  Miriam Connor, John Bauer, and Chris Manning. 2014.
\newblock \href
  {http://www.lrec-conf.org/proceedings/lrec2014/pdf/1089_Paper.pdf} {A gold
  standard dependency corpus for {E}nglish}.
\newblock In \emph{Proceedings of the Ninth International Conference on
  Language Resources and Evaluation ({LREC}'14)}, pages 2897--2904, Reykjavik,
  Iceland. European Language Resources Association (ELRA).

\bibitem[{Sinha et~al.(2021)Sinha, Jia, Hupkes, Pineau, Williams, and
  Kiela}]{sinha-etal-2021-masked}
Koustuv Sinha, Robin Jia, Dieuwke Hupkes, Joelle Pineau, Adina Williams, and
  Douwe Kiela. 2021.
\newblock \href {https://doi.org/10.18653/v1/2021.emnlp-main.230} {Masked
  language modeling and the distributional hypothesis: Order word matters
  pre-training for little}.
\newblock In \emph{Proceedings of the 2021 Conference on Empirical Methods in
  Natural Language Processing}, pages 2888--2913, Online and Punta Cana,
  Dominican Republic. Association for Computational Linguistics.

\bibitem[{Sprouse et~al.(2018)Sprouse, Yankama, Indurkhya, Fong, and
  Berwick}]{sprouse-etal-2018-colorless}
Jon Sprouse, Beracah Yankama, Sagar Indurkhya, Sandiway Fong, and Robert~C.
  Berwick. 2018.
\newblock \href {https://doi.org/doi:10.1515/tlr-2018-0005} {Colorless green
  ideas do sleep furiously: gradient acceptability and the nature of the
  grammar}.
\newblock \emph{The Linguistic Review}, 35(3):575--599.

\bibitem[{Tenney et~al.(2019)Tenney, Das, and Pavlick}]{tenney-etal-2019-bert}
Ian Tenney, Dipanjan Das, and Ellie Pavlick. 2019.
\newblock \href {https://doi.org/10.18653/v1/P19-1452} {{BERT} rediscovers the
  classical {NLP} pipeline}.
\newblock In \emph{Proceedings of the 57th Annual Meeting of the Association
  for Computational Linguistics}, pages 4593--4601, Florence, Italy.
  Association for Computational Linguistics.

\bibitem[{Vania et~al.(2019)Vania, Kementchedjhieva, S{\o}gaard, and
  Lopez}]{vania-etal-2019-systematic}
Clara Vania, Yova Kementchedjhieva, Anders S{\o}gaard, and Adam Lopez. 2019.
\newblock \href {https://doi.org/10.18653/v1/D19-1102} {A systematic comparison
  of methods for low-resource dependency parsing on genuinely low-resource
  languages}.
\newblock In \emph{Proceedings of the 2019 Conference on Empirical Methods in
  Natural Language Processing and the 9th International Joint Conference on
  Natural Language Processing (EMNLP-IJCNLP)}, pages 1105--1116, Hong Kong,
  China. Association for Computational Linguistics.

\bibitem[{Wang and Eisner(2016)}]{wang-eisner-2016-galactic}
Dingquan Wang and Jason Eisner. 2016.
\newblock \href {https://doi.org/10.1162/tacl_a_00113} {The galactic
  dependencies treebanks: Getting more data by synthesizing new languages}.
\newblock \emph{Transactions of the Association for Computational Linguistics},
  4:491--505.

\bibitem[{Warstadt et~al.(2020)Warstadt, Parrish, Liu, Mohananey, Peng, Wang,
  and Bowman}]{warstadt-etal-2020-blimp}
Alex Warstadt, Alicia Parrish, Haokun Liu, Anhad Mohananey, Wei Peng, Sheng-Fu
  Wang, and Samuel~R. Bowman. 2020.
\newblock \href {https://aclanthology.org/2020.scil-1.47} {{BL}i{MP}: A
  benchmark of linguistic minimal pairs for {E}nglish}.
\newblock In \emph{Proceedings of the Society for Computation in Linguistics
  2020}, pages 409--410, New York, New York. Association for Computational
  Linguistics.

\bibitem[{Ylonen(2022)}]{ylonen-2022-wiktextract}
Tatu Ylonen. 2022.
\newblock \href {https://aclanthology.org/2022.lrec-1.140} {Wiktextract:
  {W}iktionary as machine-readable structured data}.
\newblock In \emph{Proceedings of the Thirteenth Language Resources and
  Evaluation Conference}, pages 1317--1325, Marseille, France. European
  Language Resources Association.

\bibitem[{Zerrouki(2010)}]{zerrouki-2010-pyarabic}
Taha Zerrouki. 2010.
\newblock \href {https://pypi.python.org/pypi/pyarabic} {Pyarabic, an arabic
  language library for python}.
\newblock \url{https://pypi.python.org/pypi/pyarabic}.

\end{thebibliography}

\appendix

\section{Nonce treebanks}\label{app:nonce}

\subsection{Languages and resources}\label{app:nonce-langs} 

Table \ref{tab:nonce-languages} summarizes the languages for which \ourdata treebanks are generated. 
In general, we use the UD test sets for evaluation, and also sample replacement candidates from the UD test sets to avoid data leakage. 
For all experiments, we used the UD release 2.10.

\begin{table*}
  \centering\footnotesize
    \begin{tabular}{l|llll}\hline
  & \textbf{Family, Genus} & \textbf{Writing system} & \textbf{UD treebank} & \textbf{UD}$_{train}$ tokens \\\hline
  Arabic & Afro-Asiatic, Semitic & Arabic  & PADT, \citep{hajic-etal-2009-praguearabic} & 224K \\
  English & IE, Germanic & Latin & EWT, \citep{silveira-etal-2014-gold} & 205K    \\
  French & IE, Romance & Latin & GSD, \citep{guillaume-etal-2019-conversion} & 355K \\
  German & IE, Germanic & Latin & HDT, \citep{borges-volker-etal-2019-hdt} & 2,754K\\
  Russian & IE, Slavic & Cyrillic & SynTagRus, \citep{droganova-etal-2018-data} & 1,206K \\\hline
    \end{tabular}
  \caption[Nonce languages summary]{Summary of the languages for which nonce data is generated. IE = Indo-European. Family and Genus according to \citet{wals}.}
  \label{tab:nonce-languages}
  \end{table*}

\subsection{Language-specific rules}

For consistency, we implement the general rule in languages written in Latin script that the replacement has to be capitalized if the replaced word was capitalized. 

\paragraph{Arabic} No language-specific rules are applied, except for the removal of diacritics to increase the number of possible replacements.  

\paragraph{German} Adjective suffixes are depending on several features, however, not all of these features are consistently annotated in the available resources. 
Concretely, the forms are inflected for Case, Number, Genus, Degree of comparison, and Determinacy, leading to large inflection paradigms. 
For this reason, we implement the rule that for adjectives ending in %
\nl{-e, -em, -en, -er, -es}, the replaced adjective has to have the same ending. 

\paragraph{English} A word list is compiled from wiktionary using wiktextract \citep{ylonen-2022-wiktextract} to determine words starting with consonants or vowels. 
This list is used to determine the correct indefinite article for English nouns: 
The choice of \nl{a} or \nl{an} depends on the following word, which can be a replaced content word (e.g., an adjective or noun). 
The determiner is adjusted if the first sound of the following word changes from a vowel to a consonant or vice versa. 
For example, when replacing \nl{apple} in \nl{an apple} with \nl{bicyle}, the result is \nl{a bicycle} and not \nl{*an bicycle}. 

\paragraph{French} In French, we implement a similar rule as for English determiners. 
Concretely, we replace \nl{le/la/de/} with \nl{l'/d'} and vice versa if the following word starts with a vowel, consonant, or aspirated \nl{h}, respectively. 
The pronunciation of the following word is determined from wiktionary using wiktextract \citep{ylonen-2022-wiktextract}. 
Furthermore, French adjectives fall into three classes: 
A fixed set of adjectives precedes the noun. 
The majority of adjectives follow the noun. 
A small set of adjectives can appear in both positions. 
For instance, \nl{grande maison} -- `big house' and \nl{voiture rouge} -- `red car' are correct, but \nl{*maison grande} and \nl{*rouge maison} are not. 
The syntactic context as defined in generating nonce data does not capture the adjective classes. 
Thus, we replace adjectives preceding (following) the head only with adjectives that also precede (follow) the head in the UD treebank. 

\paragraph{Russian}
The rich case system in Russian poses a challenge for the algorithm as defined above. Concretely, the case of dependents is not part of the syntactic context. 
Contrary to the other languages in our sample, the case of an \texttt{obj}ect is determined by the verb, and both accusative and genetive case are possible and frequent cases for these objects. 
This means that if a verb with an accusative object is replaced with a verb with a genitive object, a case mismatch is introduced. 
Since Russian nevertheless retains high probing performance (Sec.~\ref{sec:nonce-probing}), we did not explicitly address this problem.

\subsection{Examples}\label{app:nonce-exs}

\begin{figure*}[!h]\centering
    \includegraphics[scale=0.6]{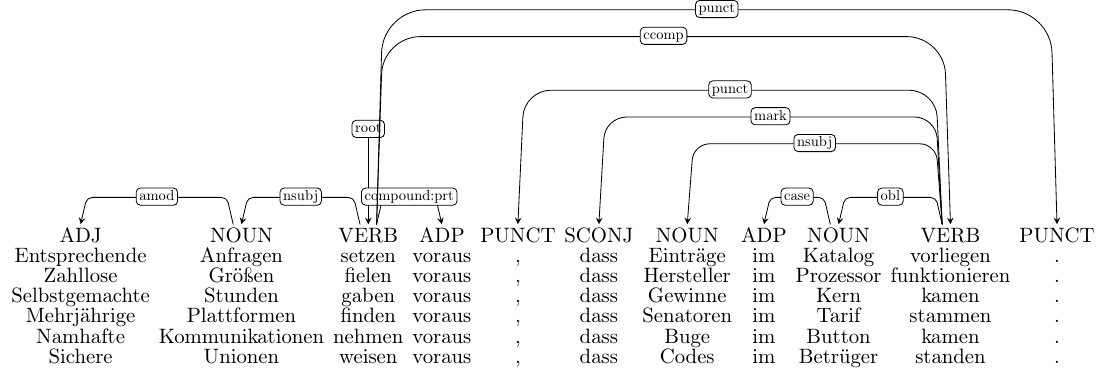}

    \includegraphics[scale=0.65]{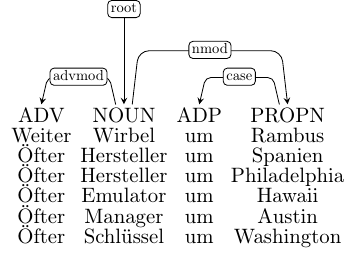}
\hspace{1cm}\includegraphics[scale=0.65]{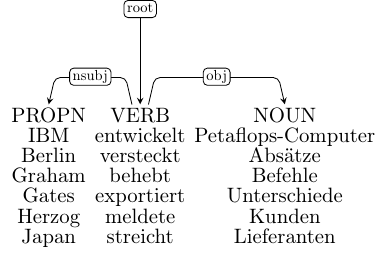}
    \caption{German \ourdata examples.}
    \label{fig:de-exs}
\end{figure*}

\begin{figure*}[!h]\centering
    \includegraphics[scale=0.6]{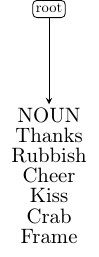}
    \hspace{1cm} \includegraphics[scale=0.6]{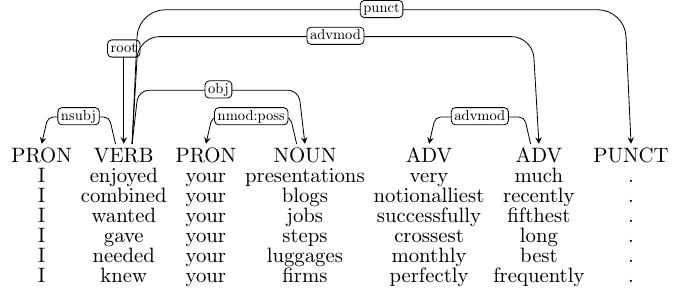}

    \includegraphics[scale=0.6]{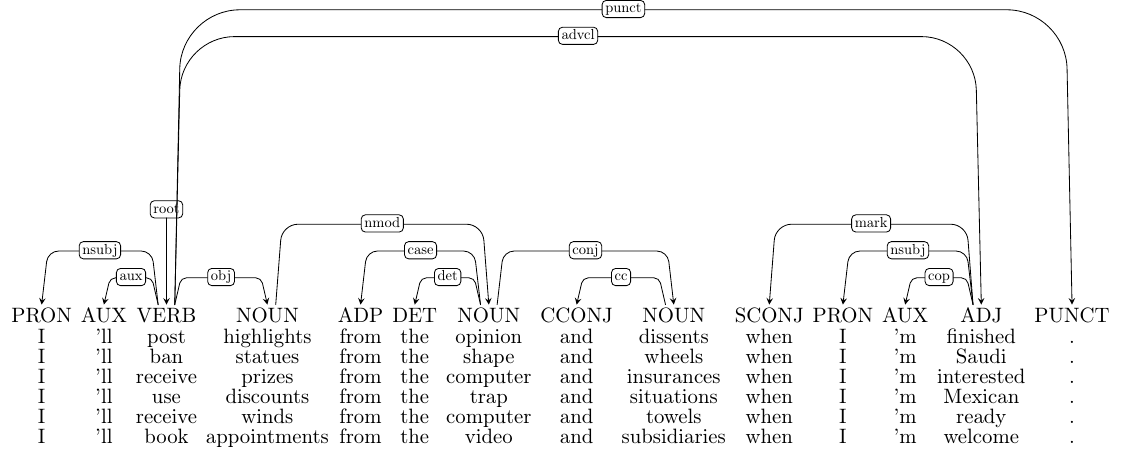}
    \caption{English \ourdata examples.}
    \label{fig:en-exs}
\end{figure*}

\begin{figure*}[!h]\centering
    \includegraphics[scale=0.65]{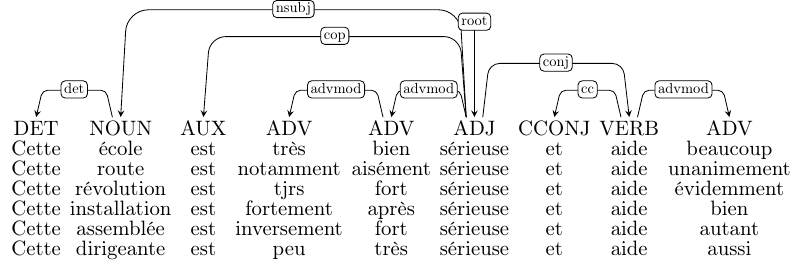}
    \includegraphics[scale=0.65]{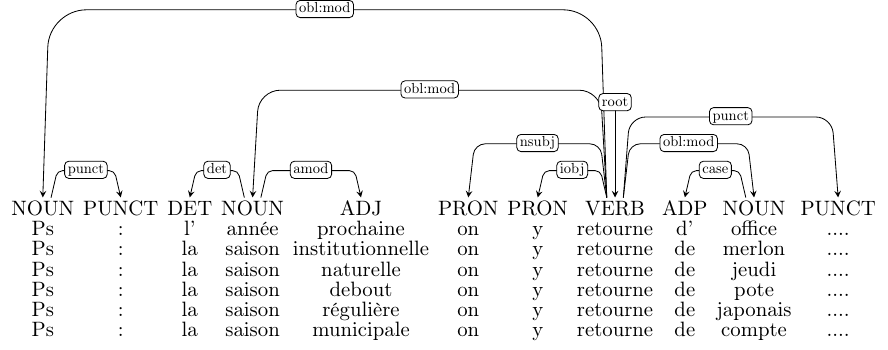}
    \includegraphics[scale=0.65]{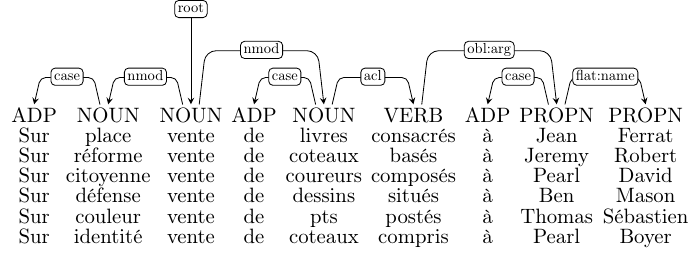}
    \caption{French \ourdata examples.}
    \label{fig:fr-exs}
\end{figure*}

\begin{figure*}[!h] \centering
    \includegraphics[scale=0.65]{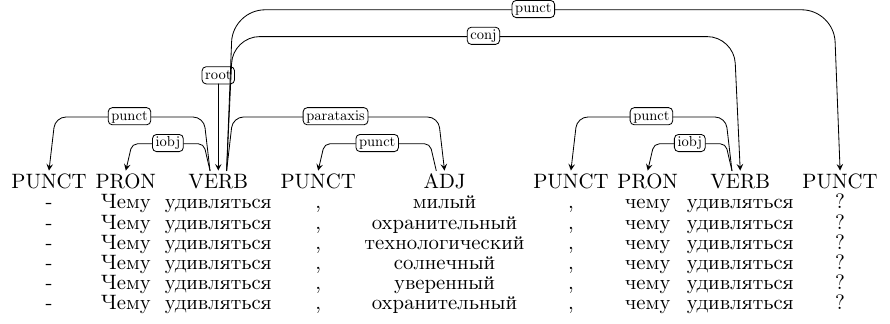}
    \includegraphics[scale=0.65]{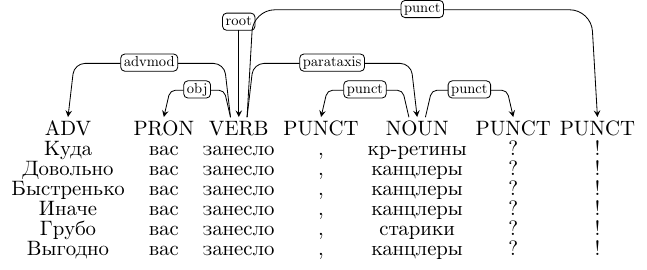}
    \includegraphics[scale=0.65]{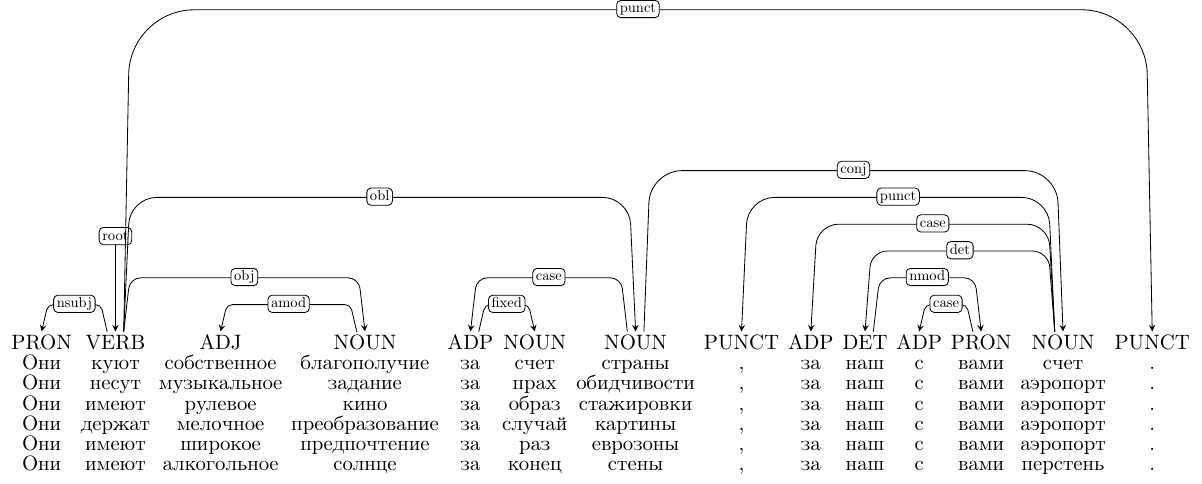}
    \caption{Russian \ourdata examples.}
    \label{fig:ru-exs}
\end{figure*}

For illustration purposes, we present relatively short sentences with multiple nonce versions in Figures 
\ref{fig:ar-exs} (ar), 
\ref{fig:de-exs} (de), 
\ref{fig:en-exs} (en), 
\ref{fig:fr-exs} (fr), and
\ref{fig:ru-exs} (ru). 

\begin{figure*}[!h]\centering
    \scalebox{.8}{
    \begin{dependency}
        \begin{deptext}
        NOUN \& NOUN \& NOUN \& X \& X \& PUNCT \& NOUN \& ADP \& NOUN \\
        alsayid \& almuhandis \& alduktur \& ghasaan \& tayaarat \& - \& waziran \& lilsinaeat \& lilsinaea \\
        alsayid \& khabir \& najah \& ghasaan \& tayaarat \& - \& janub \& lilsinaeat \& batn \\
        alsayid \& tawasue \& shakhs \& ghasaan \& tayaarat \& - \& muhamad \& lilsinaeat \& hal \\
        alsayid \& hal \& eamil \& ghasaan \& tayaarat \& - \& muhamad \& lilsinaeat \& dabit \\
        alsayid \& sahm \& niqash \& ghasaan \& tayaarat \& - \& janub \& lilsinaeat \& qamh \\
        alsayid \& yawm \& mil \& ghasaan \& tayaarat \& - \& janub \& lilsinaeat \& ghad \\
        \end{deptext} 
        
        \deproot{1}{root}
        \depedge{1}{2}{nmod}
        \depedge{1}{3}{nmod}
        \depedge{5}{4}{nmod}
        \depedge{1}{5}{nmod}
        \depedge{1}{6}{punct}
        \depedge{1}{7}{dep}
        \depedge{9}{8}{case}
        \depedge{7}{9}{nmod}
        
        \end{dependency}
    }
        \scalebox{.8}{
        \begin{dependency}
            \begin{deptext}
            NUM \& SYM \& NOUN \& ADP \& NOUN \& NOUN \& ADP \& X \\
            90 \& \% \& ziadat \& fi \& sadirat \& almansujat \& 'iilaa \& turkia \\
            90 \& \% \& masar \& fi \& tawasue \& waqud \& 'iilaa \& turkia \\
            90 \& \% \& aism \& fi \& fariq \& daght \& 'iilaa \& turkia \\
            90 \& \% \& hukm \& fi \& sijn \& eali \& 'iilaa \& turkia \\
            90 \& \% \& eumar \& fi \& murashah \& fibrayir \& 'iilaa \& turkia \\
            90 \& \% \& bad' \& fi \& fard \& shakhs \& 'iilaa \& turkia \\
            \end{deptext}
            
            \deproot{1}{root}
            \depedge{1}{2}{nmod}
            \depedge{1}{3}{nsubj}
            \depedge{5}{4}{case}
            \depedge{3}{5}{nmod}
            \depedge{5}{6}{nmod}
            \depedge{8}{7}{case}
            \depedge{5}{8}{nmod}
            
            \end{dependency}
        }

\caption{Arabic \ourdata examples (in transliteration).}
    \label{fig:ar-exs}
\end{figure*}

\subsection{Replacement statistics}

\begin{table}
\scalebox{0.9}{
\begin{tabular}{l|rrrrrr}
  \toprule
  & \texttt{NOUN} &  \texttt{PROPN} & \texttt{ADJ} &  \texttt{ADV} &  \texttt{VERB} & total \\
  \midrule
  ar &            0.93 &             0.77 &           0.93 &           0.74 &            0.66 &              0.47 \\
  de &            0.96 &             0.93 &           0.68 &           0.92 &            0.61 &              0.41 \\
  en &            0.85 &             0.84 &           0.86 &           0.83 &            0.59 &              0.38 \\
  fr &            0.83 &             0.69 &           0.88 &           0.92 &            0.50 &              0.36 \\
  ru &            0.84 &             0.70 &           0.80 &           0.89 &            0.46 &              0.39 \\
  \bottomrule
\end{tabular}
}
\caption{Ratio of replaced words per POS in the test set. The column "total" shows the replacement ratio over all POS, including function words and punctuation.}
\label{tab:repl-stats}
\end{table}

Tab.~\ref{tab:repl-stats} shows the ratio of replaced words per POS in the test set. 
Of the 25 ratios (5 languages, 5 POS tags), 6 are above 0.9 and 14 are above 0.8, indicating that the algorithm is able to find replacements for the vast majority of content words. 
The replacement ratio is lowest for verbs, for two reasons: 
First, verbs are generally less frequent than e.g. nouns.
Second, verbs frequently have a relatively complex argument structure, which leads to a large number of syntactic contexts and an increased frequency of syntactic contexts that are not shared with other verbs.

\section{List of Models}\label{app:lms}

mGPT \cite[huggingface id: \texttt{ai-forever/mGPT}]{shliazhko-etal-2022-mgpt} and 
mBERT \cite[\texttt{bert-base-multilingual-cased}]{devlin-etal-2019-bert} are used for all experiments. 
English scoring and 
probing experiments are additionally conducted with 
RoBERTa \cite[\texttt{roberta-base}]{liu-etal-2019-roberta} and
GPT-2 \cite[\texttt{gpt2}]{radford-etal-2019-gpt2}. 
Arabic scoring experiments are additionally conducted with
AraGPT2 \cite[\texttt{aubmindlab/aragpt2-base}]{antoun-etal-2021-aragpt2} and
AraBERT \cite[\texttt{aubmindlab/bert-base-arabertv2}]{antoun-etal-2020-arabert}.

\section{Data preprocessing}

We apply the following preprocessing steps to the UD treebanks after creating \ourdata splits. 
To allow a fair comparison between MLMs and ALMs, we filter out sentences that are shorter than 4 words from the data in all experiments. 
On short sentences, ALMs have a disadvantage because their per-token score increases over the sentence given increasing context length, while the basic masked language modeling task assumes access to the same amount of context for each token. 
For \ourdata in Arabic, we remove diacritics with the \texttt{PyArabic} library \citep{zerrouki-2010-pyarabic}. 
The Arabic morphological lexicon is lemmatized with Farasa \citep{darwish-mubarak-2016-farasa}.

\section{Lexical Diversity of Nonce data}\label{app:ttr}

\begin{table}[h!]
    \centering
    \begin{tabular}{l|rr}
      \toprule
    & \multicolumn{2}{c}{TTR} \\
    Language & orig. & nonce  \\
    \midrule
    ar & .0094 & .0077 \\
    de & .0054 & .0043 \\
    en & .0571 & .0463 \\
    fr & .0441 & .0297 \\
    ru & .0052 & .0044 \\
    \bottomrule
  \end{tabular}
    \caption{Type-Token Ratio of underlying UD treebanks and \ourdata versions.} %
    \label{tab:intrinsic-results}
\end{table}
\paragraph{Model-Independent Comparison: Type Token Ratio (TTR)} 
is a widely used measure of lexical diversity. where a larger value indicates that less tokens are used repeatedly \citep{koehler-2003-typetoken}. 
Apart from the segmentation model (in our case the mBERT tokenizer), TTR focuses on capturing the distribution of the data relatively independent from large computational or methodological overhead. 
The TTR (when tokenized with mBERT) in all languages is shown in Tab.~\ref{tab:intrinsic-results}.

\section{Scoring Experiments}

\subsection{Example for Scoring Functions}\label{app:scoring-ex}

An example for how all scoring functions are computed for the phrase \nl{accordeon player} is displayed in Tab.~\ref{tab:score-fn-examples}.

\begin{table}[!h]\centering\footnotesize
  \begin{tabular}{l|lll}
      \toprule
      {} & accord & \#\#eon & player \\
      \midrule
       & ? & \_ & \_ \\
       $PPL$         & accord & ? & \_ \\
                   & accord & \#\#eon & ? \\\hline
       & ? & \#\#eon & player \\
       $PPPL$           & accord & ? & player \\
                    & accord & \#\#eon & ? \\\hline
       & ? & \_ & player \\
       $PPPL_{l2r}$           & accord & ? & player \\
                   & accord & \#\#eon & ? \\
      \bottomrule
  \end{tabular}
  \caption{The token for which a probability is recorded is marked with "?". 
  Tokens that are masked out or not available from context are marked with "\_".}
  \label{tab:score-fn-examples}
\end{table}

\subsection{Scoring Experiment}\label{app:score-ratios}

In Tab.~\ref{tab:score-ratios-alm}, we show the ratios of scoring functions for multilingual models in German, French and Russian (complementing the English and Arabic results in Tab.~\ref{fig:score-ratio-boxplots}). 
In Tab.~\ref{tab:score-ratios-monoling}, we show the ratios of scoring functions for monolingual models in Arabic and English. 

\begin{table}[h]
\scalebox{.6}{
\begin{tabular}{lccc}
\toprule
{} & $r_{PPPL}$ & $r_{PPPL_{l2r}}$ & $r_{PPL}$ \\
\midrule
de & \includegraphics[width=.2\textwidth]{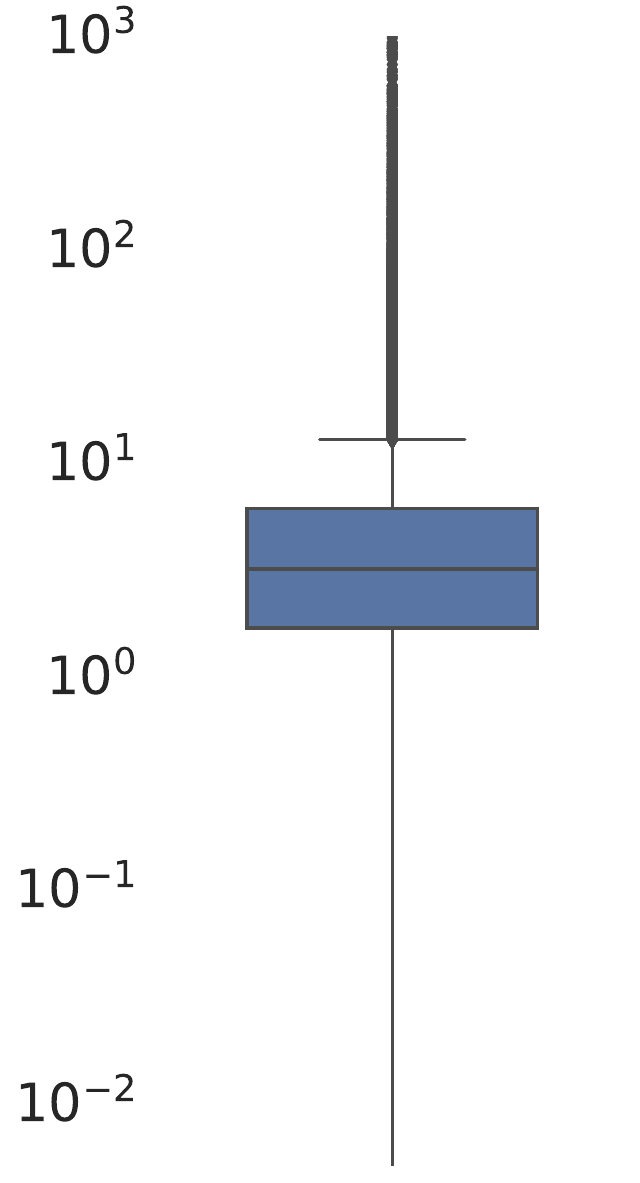} & \includegraphics[width=.2\textwidth]{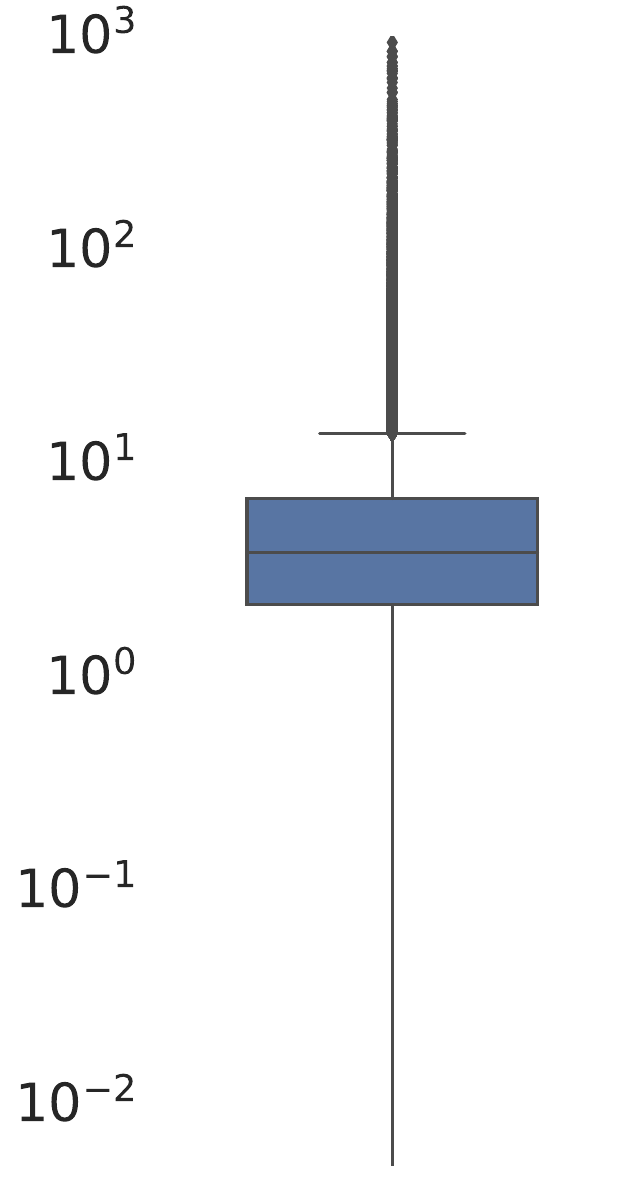} & \includegraphics[width=.2\textwidth]{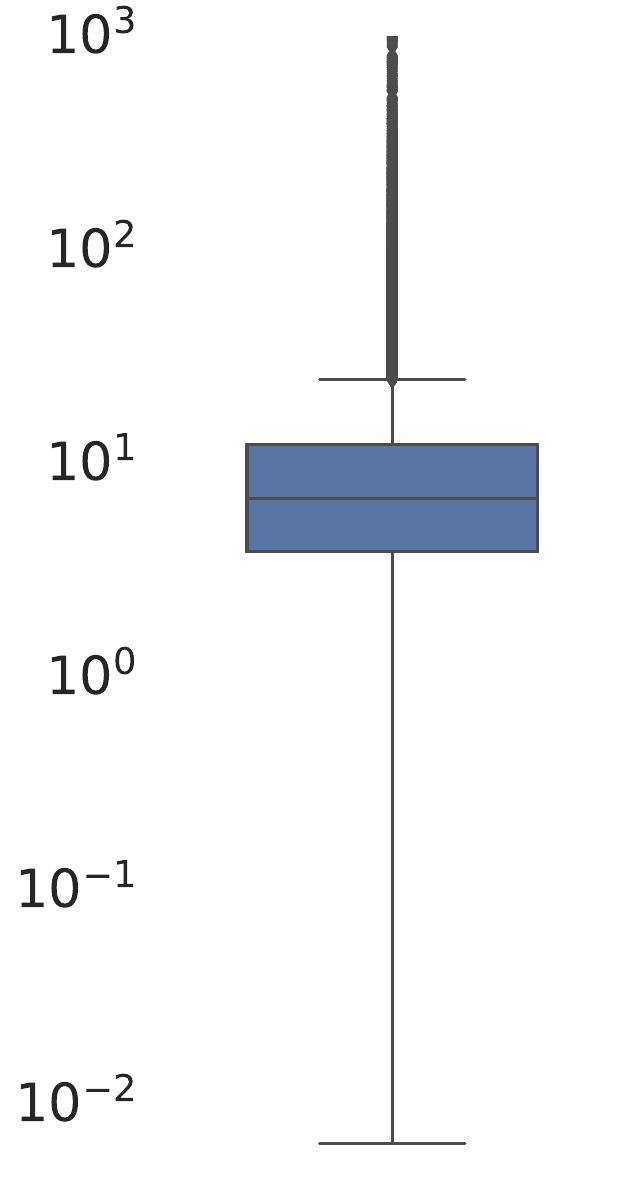} \\
fr & \includegraphics[width=.2\textwidth]{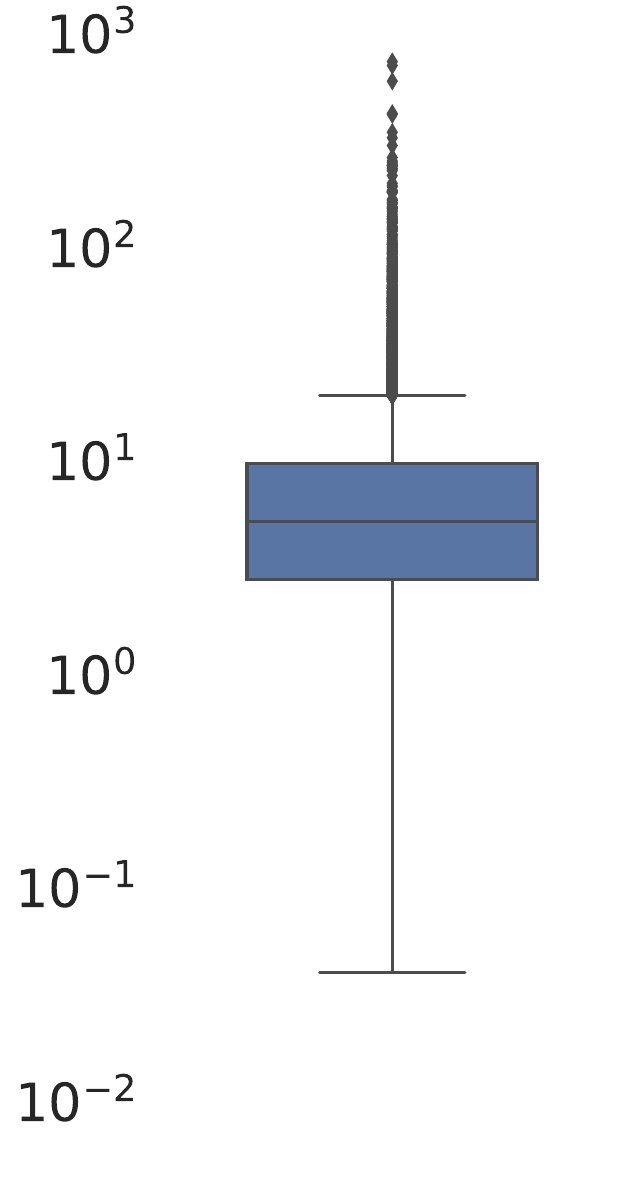} & \includegraphics[width=.2\textwidth]{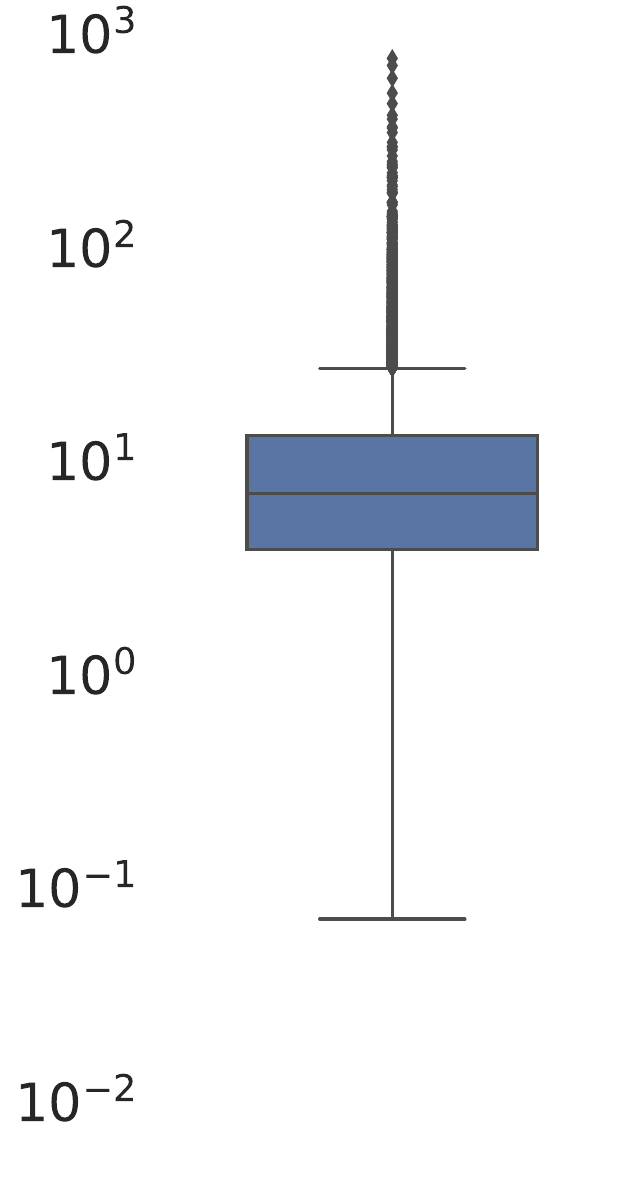} & \includegraphics[width=.2\textwidth]{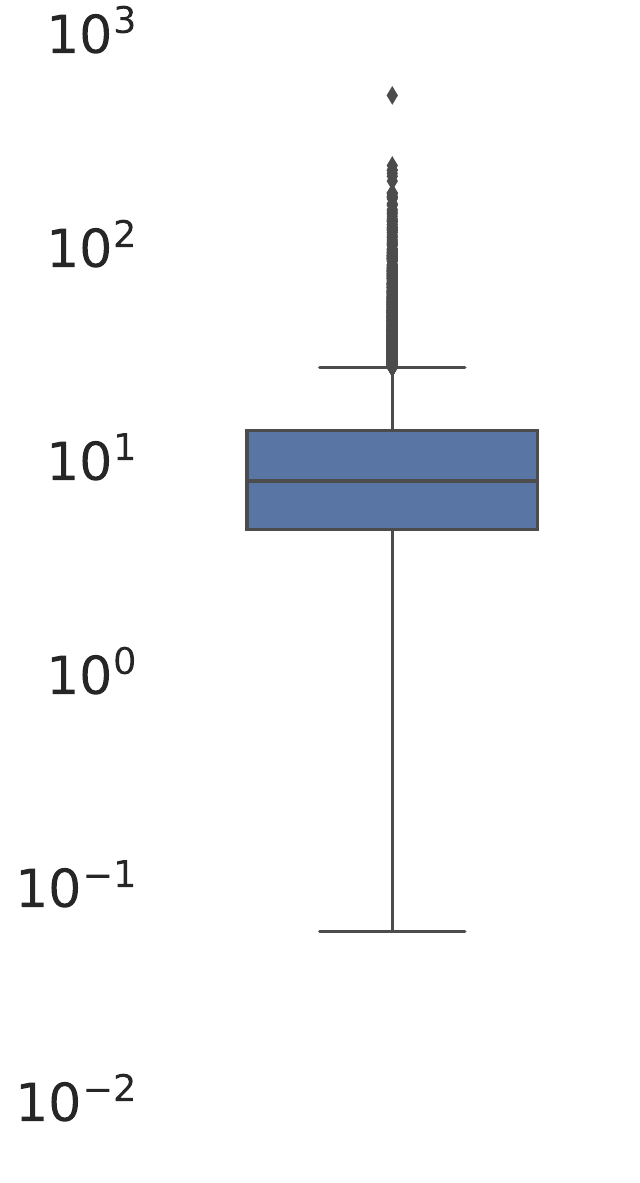} \\
ru & \includegraphics[width=.2\textwidth]{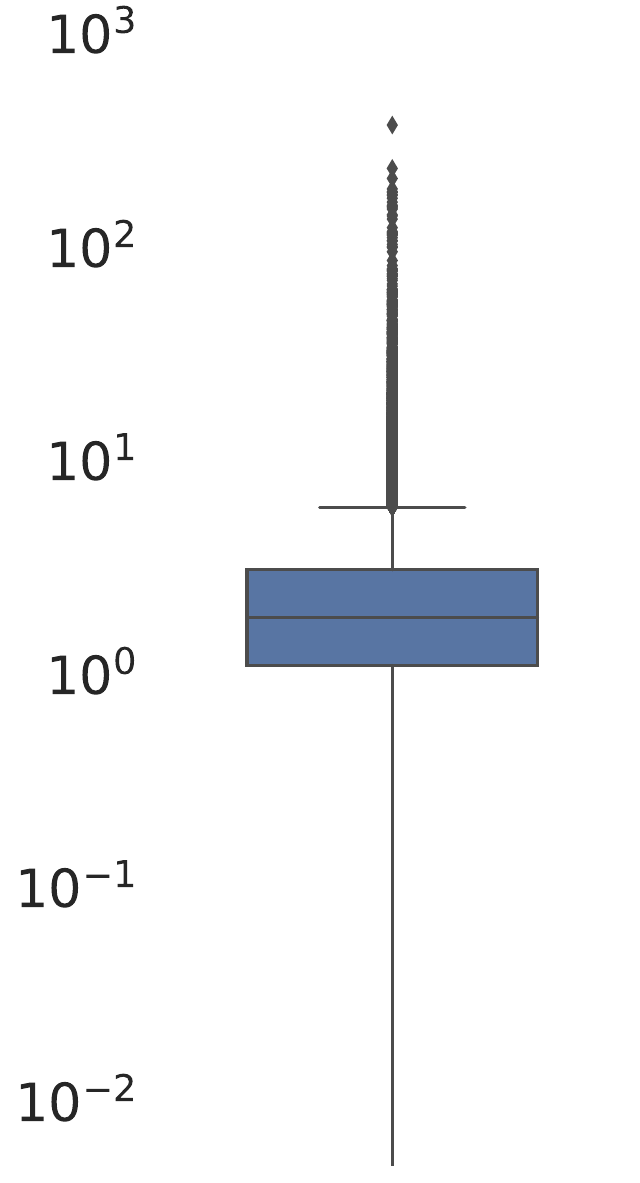} & \includegraphics[width=.2\textwidth]{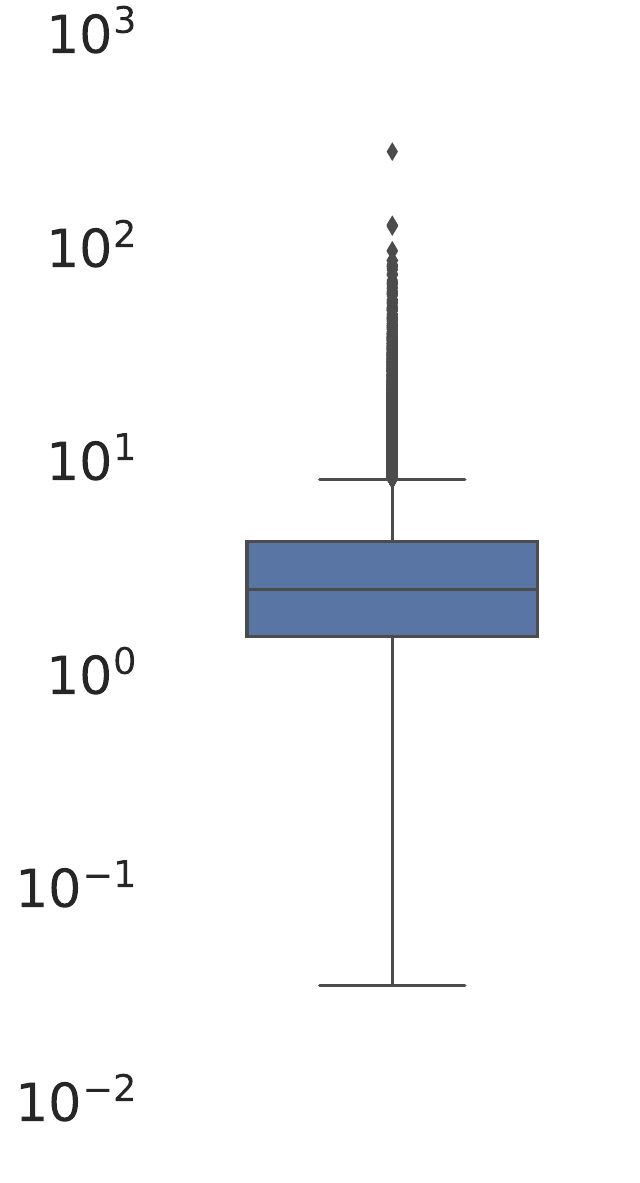} & \includegraphics[width=.2\textwidth]{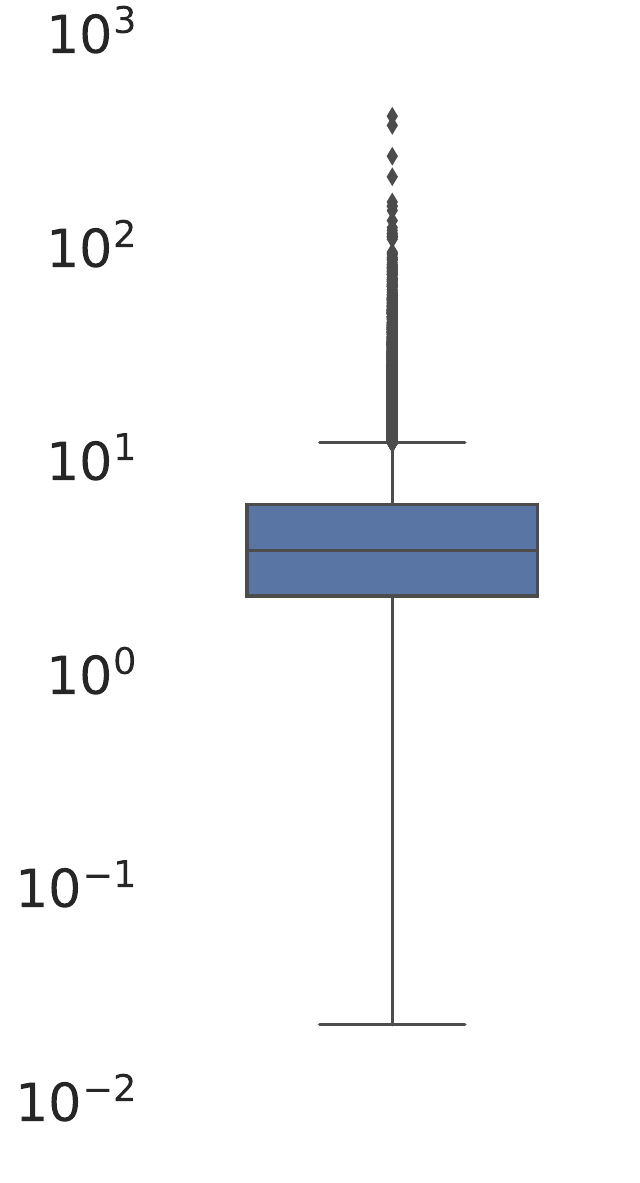} \\
\bottomrule
\end{tabular}
}
\caption{Scoring function ratios for multilingual models.}
\label{tab:score-ratios-alm}
\end{table}

\begin{table}
\scalebox{.6}{
\begin{tabular}{lccc}
  \toprule
  {} & $r_{PPPL}$ & $r_{PPPL_{l2r}}$ & $r_{PPL}$ \\
  \midrule
ar & \includegraphics[width=.2\textwidth]{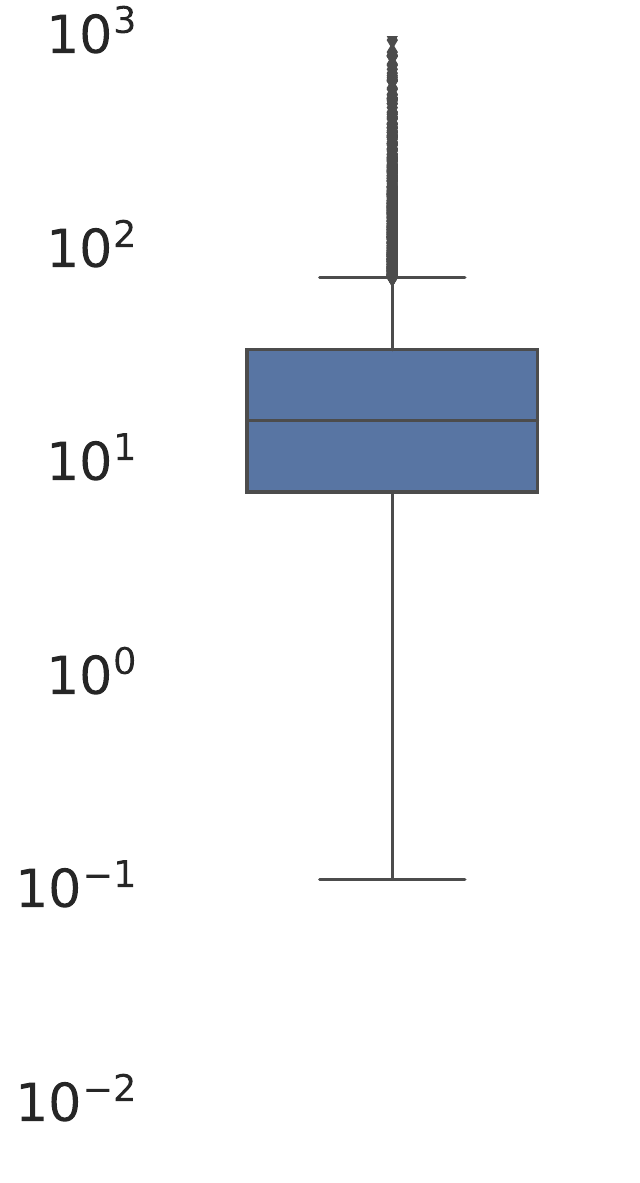} & \includegraphics[width=.2\textwidth]{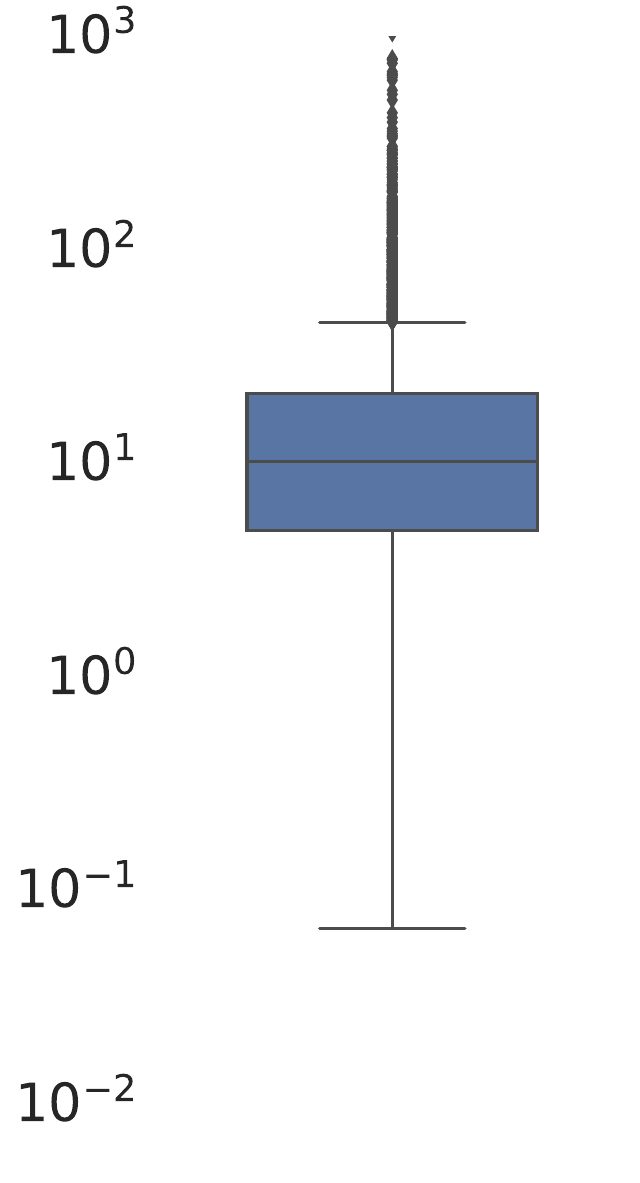} & \includegraphics[width=.2\textwidth]{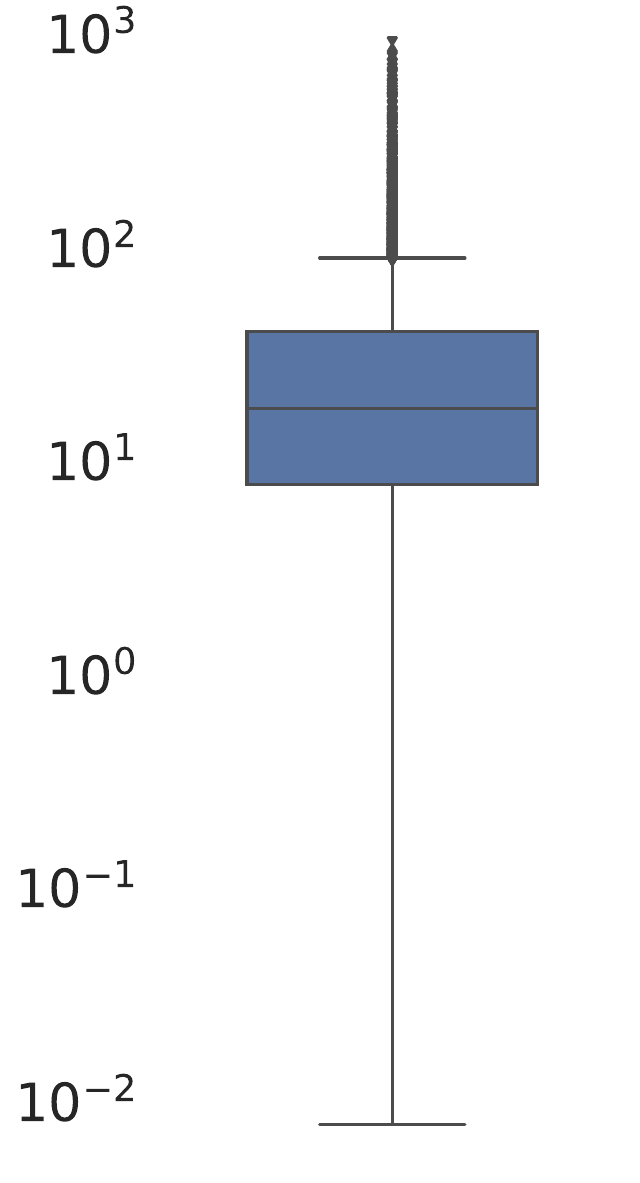} \\
en & \includegraphics[width=.2\textwidth]{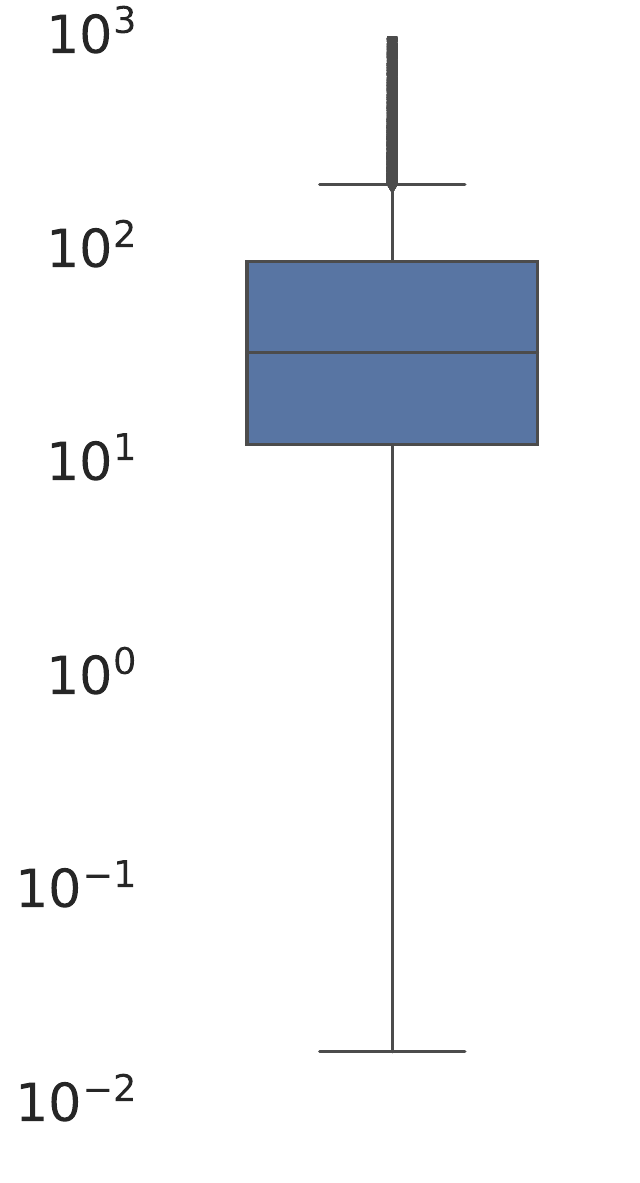} & \includegraphics[width=.2\textwidth]{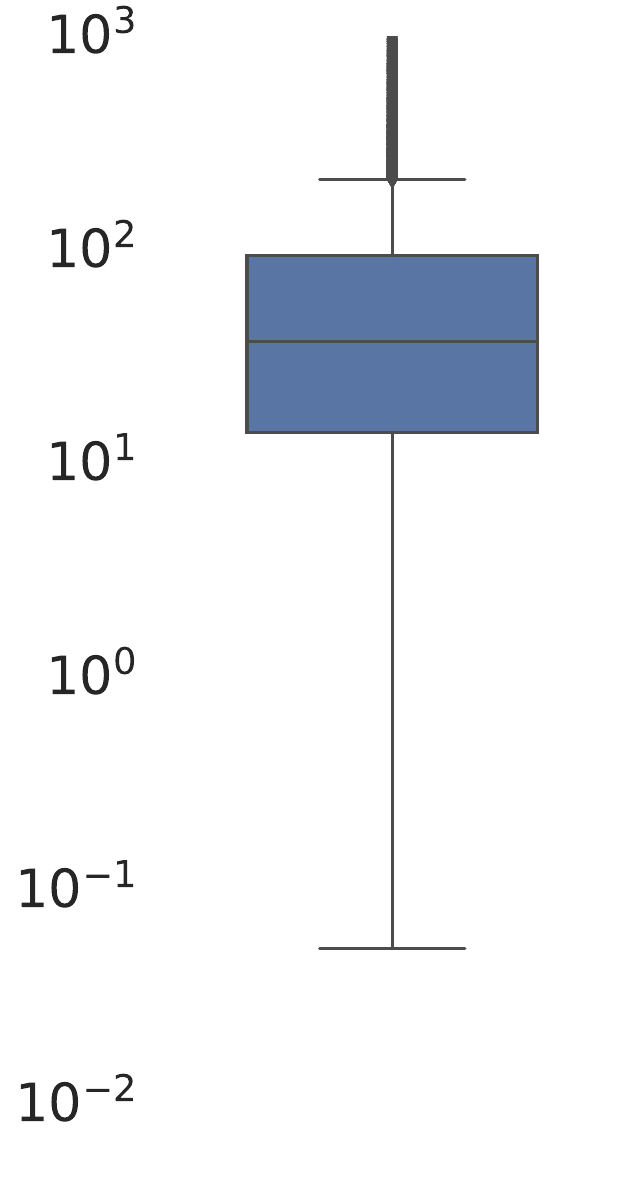} & \includegraphics[width=.2\textwidth]{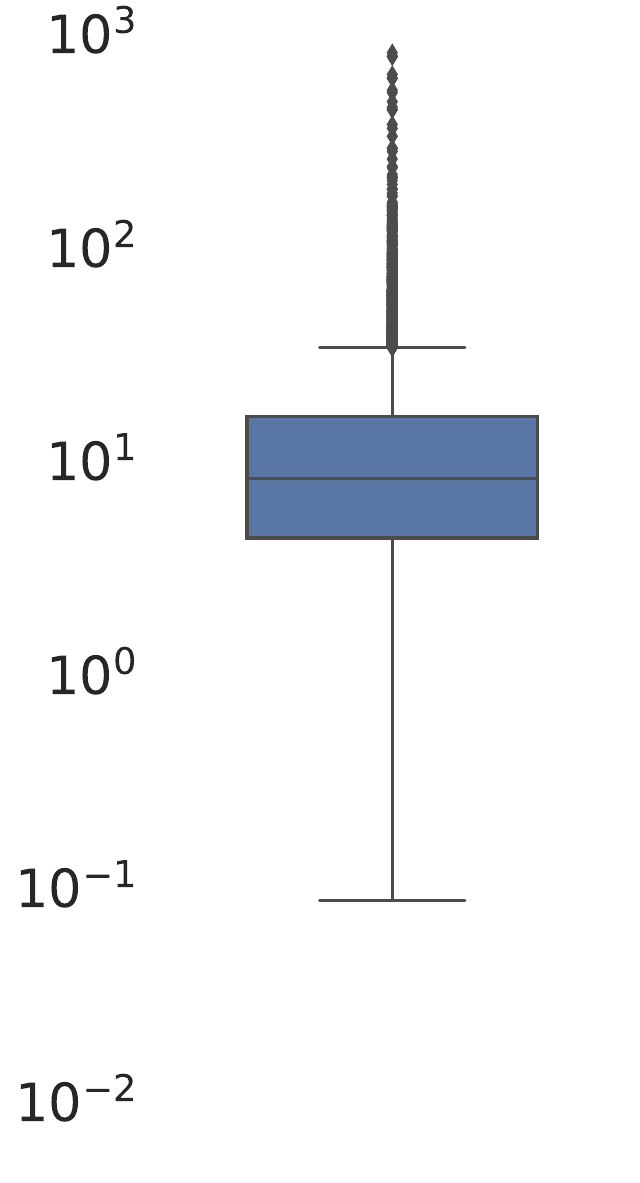} \\
  \bottomrule
\end{tabular}
}
\caption{Scoring function ratios for monolingual models (Arabic and English).}
\label{tab:score-ratios-monoling}
\end{table}

\subsection{Scoring outlier analysis}

For English, we investigate the sentence pairs with the highest and lowest scoring function ratios. 
The sentences with the highest ratio are shown in Table \ref{tab:sents-ratio-max}.
Sentences with very high ratios are all relatively short. 
The large difference in ratios between scores assigned by mBERT and mGPT indicate that these original sentences could be part of the pre-training data for mGPT but not mBERT - the former is overfitting to the original sentence. 
Also, these sentences contain uppercase sentences (which might affect tokenization), and almost all original sentences contain well-known named entities or common phrases that are disrupted in the nonce data (\nl{U.S. Senate Committee}, \nl{AS FAR AS POSSIBLE}, etc.)

The sentences with the lowest ratio are shown in Table \ref{tab:sents-ratio-min}.
The examples include typos in the original data (\nl{ling} instead of \nl{long}), 
an example in which the original seems ungrammatical (6), and examples in which only single words are replaced by other forms that are presumably more frequent (\nl{cover - go}, \nl{Per - Google}, \dots)

\begin{table*}[h]\centering
  \scalebox{.66}{
  \begin{tabular}{llrrr}
    \toprule
                                                 orig. &                                            nonce &  $r_{PPPL}$ &  $r_{PPPL_l2r}$ &  $r_{PPL}$ \\
    \midrule
    NORTH CAROLINA RELIGIOUS COALITION FOR MARRIAGE... &              D' Vernon ugly Coalition For El Dong &      388.97 &          176.08 &        3320.79 \\
               U.S. Senate Committee on Appropriations &                     Can Bay Committee on Williss' &    35734.78 &        16130.36 &        1027.64 \\
          SHE KNOWS GREAT FOOD AND DINING EXPERIENCES. & She solicits Swedish election And rabbit outlets. &        8.67 &           16.67 &         968.43 \\
                       BEST CHINESE RESTAURANT EVER!!! &           Truest corrupt Restaurant definitely!!! &        4.57 &            7.44 &         855.44 \\
                         THEY ARE VERY RUDE AND NASTY. &              They Are allegedly happy And hidden. &       80.31 &           38.98 &         797.19 \\
                my new OLYMPUS X940 DIGITAL CAMERA...? &         My inhuman April Port wireless Camera...? &       10.03 &            5.29 &         759.14 \\
                         STAY AWAY AS FAR AS POSSIBLE. &      Move wrongest currently low As satisfactory. &       31.35 &           28.88 &         608.69 \\
               Super Pet Silent Spinner Exercise Wheel &   Standard Fidelity alleged Alliance Prince Wheel &      257.01 &           74.85 &         538.77 \\
                      CHERNOBYL ACCIDENT: TEN YEARS ON &            Greenfield diplomacy: Ten Years partly &       10.86 &           17.24 &         511.29 \\
    \bottomrule
    \end{tabular}
  }
    \caption{Selection of sentences with $r_{PPL}$ above 500. Data is printed verbatim, including casing, punctuation, etc.}
    \label{tab:sents-ratio-max}
\end{table*}

\begin{table*}[h]
  \centering
  \scalebox{.8}{
    \begin{tabular}{llrrr}
      \toprule
                                                    orig &                                              nonce &  $r_{PPPL}$ &  $r_{PPPL_l2r}$ &  $r_{PPL}$ \\
      \midrule
                        New training Centre is excellent &          Extensive vacation temperature is amazing &        0.01 &            0.32 &           0.04 \\
                           Maria Valdes superior \$62,500 &                       Todd Taylors' strong \$62,500 &        0.47 &            0.24 &           0.11 \\
                     Gold award parts excellence, metro. &              Hoodie leader coffees tank, Michelle. &        1.60 &            2.17 &           0.14 \\
                    Talked to Craig and the Court today. &          Talked to Emily and the viewer yesterday. &        0.05 &            0.43 &           0.16 \\
                              oh god is there an agenda. &                     Oh god is there an antibiotic. &        0.22 &            0.20 &           0.18 \\
                       Broke out the activities of 1179. &                   Checked out the spirits of 1179. &        0.79 &            0.69 &           0.21 \\
            Shanna, I spoke with Per tonight about this. &     Kim, I spoke with Google yesterday about this. &        0.30 &            0.30 &           0.24 \\
                              George Bush: Money manager &                   Roger's Dick's: garage therapist &        0.04 &            0.13 &           0.25 \\
        Is it possible to shoot lazers out of your Wang? &      Is it great to see e-mails out of your house? &        0.08 &            0.06 &           0.28 \\
                     if they hiss, they are not playing. &                 If they are, they are not running. &        0.62 &            0.49 &           0.34 \\
                    Even though you are expensive. &                Certainly though you are expensive. &        3.74 &            5.69 &           0.39 \\
          See you all there - this is ling overdue &              See you all there - this is here deep &        1.19 &            0.56 &           0.44 \\
                       Can you cover for me today? &                        Can you go for me tomorrow? &        0.18 &            0.43 &           0.46 \\
               Brazil we have current data already &          Danny we have ambitious tongue feverishly &        0.08 &            0.82 &           0.47 \\
                     Your cat will adjust quickly. &        Your health will starve inappropriateliest. &        0.76 &            2.27 &           0.48 \\
\bottomrule
\end{tabular}
  }
\caption{Selection of sentences with $r_{PPL} < 0.5$. Data is printed verbatim, including casing, punctuation, etc. Data with such low ratios also included longer sentences, which are omitted for space reasons.}
\label{tab:sents-ratio-min}
\end{table*}

\section{Probing Experiments}\label{app:probing}

\begin{figure*}[h]
\centering

\begin{subfigure}[b]{0.45\textwidth}
    \centering
\scalebox{0.65}{
\begin{tabular}{l|llllllll}
\toprule
& why & does & my & snake & refuse & to & eat & ? \\
\midrule
why & 0 & 2 & 3 & 2 & 1 & 3 & 2 & 2 \\
does & 2 & 0 & 3 & 2 & 1 & 3 & 2 & 2 \\
my & 3 & 3 & 0 & 1 & 2 & 4 & 3 & 3 \\
snake & 2 & 2 & 1 & 0 & 1 & 3 & 2 & 2 \\
refuse & 1 & 1 & 2 & 1 & 0 & 2 & 1 & 1 \\
to & 3 & 3 & 4 & 3 & 2 & 0 & 1 & 3 \\
eat & 2 & 2 & 3 & 2 & 1 & 1 & 0 & 2 \\
? & 2 & 2 & 3 & 2 & 1 & 3 & 2 & 0 \\
\bottomrule
\end{tabular}
}
\caption{Gold distance matrix for the example sentence.}
\label{fig:depprobe-algo-dists}
\end{subfigure}

\begin{subfigure}[b]{1\textwidth}
\centering
\scalebox{.8}{
\newcolumntype{L}[1]{>{\raggedright\let\newline\\\arraybackslash\hspace{0pt}}m{#1}}
\newcolumntype{C}[1]{>{\centering\let\newline\\\arraybackslash\hspace{0pt}}m{#1}}
\begin{tabular}{l|L{2cm}|C{7cm}|L{3cm}}
\toprule
& covered words & tree & comment \\
\midrule
1 & $\{refuse\}$ &
\begin{dependency}
    \begin{deptext}
    refuse \\
    \end{deptext}
    \deproot{1}{root}
    \end{dependency} & maximum $p(\texttt{root}|refuse)$ \\\hline
2-6 & $\{$\nl{why, does, snake, refuse, eat, ?}$\}$ & 
\begin{dependency}
\begin{deptext}
why \& does \& snake \& refuse \& eat \& ? \\
\end{deptext}
\depedge{4}{1}{advmod}
\depedge{4}{2}{aux}
\depedge{4}{3}{nsubj}
\deproot{4}{root}
\depedge{4}{5}{xcomp}
\depedge{4}{6}{punct}
\end{dependency} & all distance 1 to root \nl{refuse} \\\hline
7 & $\{$\nl{why, does, my, snake, refuse, eat, ?}$\}$ &
\begin{dependency}
\begin{deptext}
why \& does \& my \& snake \& refuse \& eat \& ? \\
\end{deptext}
\depedge{5}{1}{advmod}
\depedge{5}{2}{aux}
\depedge{4}{3}{nmod}
\depedge{5}{4}{nsubj}
\deproot{5}{root}
\depedge{5}{6}{xcomp}
\depedge{5}{7}{punct}
\end{dependency} & \nl{snake} is only word with distance 1 to \nl{my} \\\hline
8 & $\{$\nl{why, does, my, snake, refuse, to, eat, ?}$\}$ &
\input{fig/depprobe-algo-tree.tex} & \nl{to} is only word with distance 1 to \nl{eat} \\
\bottomrule
\end{tabular}
}
\caption{Step by Step example.}
\label{fig:depprobe-algo-steps}
\end{subfigure}

\caption{DepProbe tree decoding example}
\label{fig:depprobe-algo}
\end{figure*}

\subsection{DepProbe architecture}\label{app:depprobe-arch}

\citet{muller-eberstein-etal-2022-probing} presented \textbf{DepProbe}, a lightweight decoder for directed and labeled dependency trees. 
The model consists of two matrices, $L$ and $B$. 
Both transform $d_h$-dimensional word representations\footnote{For tokens consisting of multiple subwords, the word representation is the mean of the subword representations.} 
into vectors that hightlights a syntactic property of that representation. 
Assume that the LM representation of a word $w_i$ is $h_i \in \mathbb{R}^{d_h}$, and that the annotation has $l$ dependency relations.

\paragraph{Predicting dependency relations}
The matrix $L \in \mathbb{R}^{d_h \times l}$ is a linear classifier that predicts for each word the label of this word's incoming dependency edge -- the relation between the word and its head. 
Concretely, the ``probability of a word's relation $r_i$ being of class $l_k$ is given by:
\begin{equation}
    p(r_i = l_k|w_i) = softmax(Lh_i)_k
\end{equation}

'' \cite[p.~7713]{muller-eberstein-etal-2022-probing}. This model component is trained using cross-entropy loss. 

\paragraph{Predicting word distances}
The matrix $B
\in \mathbb{R}^{d_h \times b}$ predicts the dependency edges between words. 
$B$ projects the LM representations in a vector space that has less dimensions than the LM layer ($b < d_h$). 
The target vector space is informally called the syntactic subspace: %
It reflects structural information such that vector distances between words mimic the distance between words in the dependency tree. 
Concretely, when $h_i, h_j$ are the LM representations of words $w_i, w_j$, their distance $d_{B}(h_i, h_j)$
in the syntactic subspace is defined by \citet[Eq.~1]{muller-eberstein-etal-2022-probing}:
\begin{equation}
    d_{B}(h_i, h_j) = \sqrt{(Bh_i - Bh_j)^T(Bh_i - Bh_j)}
\end{equation}

This model component is trained to predict the distance between all word pairs in the dependency tree. 
Assume that the distance between two words in the dependency tree is defined as the path length between the two words, $d_P(w_i, w_j)$. 
If $s$ is a sentence of length $N+1$, the loss for optimizing $B$ is given by \citet[Eq.~2]{muller-eberstein-etal-2022-probing}:
\begin{equation}
    \mathcal{L}_{B}(s) = \frac{1}{N^2} \sum_{i=0}^{N} \sum_{j=0}^{N} | d_P(w_i, w_j) - d_{B}(h_i, h_j) |
\end{equation}

\paragraph{Constructing the dependency tree}
The outputs of the relation and distance components are combined to construct the dependency tree starting from the root. 
The word with the highest probability of being the root (as assigned by $L$) is set as the dependency tree root. 
Then, the words are iteratively added to the dependency tree, always choosing the word with the smallest distance $d_B$ to a word that is already covered. 
The dependency labels are assigned based on $L$. 
A detailed example of this process is shown in App.~\ref{app:depprobe-example}.

\subsection{DepProbe Example}\label{app:depprobe-example}

In Fig.~\ref{fig:depprobe-algo}, we present a step-by-step example for 
for reconstructing trees with DepProbe, following Alg. 1 in \cite{muller-eberstein-etal-2022-probing}. 
For simplicity, the correct distance matrix (Fig.~\ref{fig:depprobe-algo-dists}) and correct relations are taken as input to the algorithm. 

\subsection{Baselines}\label{app:baseline-scores}

\paragraph{mBERT embedding layer ($M_0$)} 
In this baseline, 
we use the output of mBERT's embedding layer as input to DepProbe. 
$M_0$ estimates how much information about the probing task is contained in the (context-insensitive) token embeddings. 

\paragraph{Random mBERT ($M_R$)}
We consider the representations of a transformer model with the same architecture as mBERT, but all parameters are randomly initialized. 
Performance of this baseline estimates how much information about dependency structures is contained in the architecture of an MLM and the DepProbe training. 

\paragraph{Random contextualization mBERT ($M_{RC}$)}
We consider a transformer model with the same architecture as mBERT, but all parameters except the embedding layer are randomly initialized. 
The performance of this baseline estimates how much information about dependency structures from linguistic context is acquired 
during pretraining of the MLM \cite{belinkov-2022-probing}.
We take the representations from layers 6 (7) as input to DepProbe's distance (relation) component.

\paragraph{Results} 
All probed models outperform the baselines by a large margin, both on original and nonce data (Tables~\ref{tab:bl-layer0}, \ref{tab:bl-randomgood}, \ref{tab:bl-randomrandom}).  
This shows that the probing task probes for contextual information (comparison to $M_0$), and that the probed information is learnt during pretraining  (comparison to $M_R$ and $M_{RC}$).
$M_0$, the embedding layer baseline, shows decent performance on the relation labeling task (Tab.~\ref{tab:bl-layer0}): 
Up to 74.6 \relacc for German, and $\Delta<5$ for all languages. 
For attachment, the absolute scores are lower and the performance drop is larger. 
This is intuitive: The embedding layer provides a basis for predicting context-insensitive information such as the most frequent dependency relation of a word type, and this information is still useful for predicting the dependency relations in \ourdata data. 
However, predicting attachment requires more contextual information, for \ourdata data even more so than for original data. 
The absolute performance and $\Delta$ of $M_{RC}$ is within 1 point to $M_0$ for all languages (Tab.~\ref{tab:bl-randomgood}).
This shows that the random contextualization does not detriment the predictability of relation labeling based on the embedding layer. 
Attachment performance is lower in terms of absolute scores for $M_{RC}$ than for $M_0$, which leads to less room for a performance drop but overall lower \las on both test sets. 
On relation labeling, $M_{RC}$ performs between 4.8 and 12.6 accuracy points better than $M_R$ on original data. 
On attachment, $M_R$ performs better than $M_{RC}$ for all languages except Arabic. 
All baseline performance drops for nonce data are relatively small. 
The most intuitive explanation for this is that on the one hand, lower overall performance leaves less room for a performance drop, and on the other hand it confirms that nonce data is created in a way in which only the co-occurence information is changed, but not the syntactic contexts in which the words generally appear (e.g. the tendency of a word to appear as a subject is preserved).

\begin{table}\centering
  \footnotesize
  \begin{tabular}{lrrrrrr}
    \toprule
    {} &  \multicolumn{2}{c}{\relacc} &  \multicolumn{2}{c}{\uas} & \multicolumn{2}{c}{\las}\\
      & orig & $\Delta$ & orig & $\Delta$ & orig & $\Delta$ \\   \midrule
    ar   &            66.1 &              4.7 &      43.3 &        5.8 &      29.9 &        5.9 \\
    de   &            74.6 &              2.8 &      49.7 &        5.5 &      40.8 &        6.1 \\
    en   &            65.4 &              1.6 &      45.6 &        3.5 &      33.7 &        3.2 \\
    fr   &            70.6 &              2.1 &      48.2 &        3.0 &      36.9 &        3.2 \\
    ru   &            71.3 &              2.0 &      46.6 &        3.3 &      35.3 &        3.3 \\
    \bottomrule
    \end{tabular}
\caption{$M_0$. Embedding layer of mBERT}\label{tab:bl-layer0}
\end{table}

\begin{table}\centering
  \footnotesize
  \begin{tabular}{lrrrrrr}
    \toprule
    {} &  \multicolumn{2}{c}{\relacc} &  \multicolumn{2}{c}{\uas} & \multicolumn{2}{c}{\las}\\
      & orig & $\Delta$ & orig & $\Delta$ & orig & $\Delta$ \\
    \midrule
    ar   &            65.4 &              4.7 &      31.0 &        4.1 &      21.9 &        4.1 \\
    de   &            75.2 &              3.2 &      46.4 &        5.3 &      38.7 &        6.1 \\
    en   &            66.4 &              1.7 &      41.9 &        3.0 &      32.0 &        2.9 \\
    fr   &            70.9 &              2.5 &      40.6 &        2.3 &      31.7 &        2.4 \\
    ru   &            71.0 &              2.0 &      42.0 &        2.7 &      32.4 &        2.8 \\
    \bottomrule
    \end{tabular}
  \caption{$M_{RC}$. Mid layers of mBERT architecture with trained embedding matrix and random transformer layers}\label{tab:bl-randomgood}
\end{table}

\begin{table}\centering
  \footnotesize
  \begin{tabular}{lrrrrrr}
    \toprule
    {} &  \multicolumn{2}{c}{\relacc} &  \multicolumn{2}{c}{\uas} & \multicolumn{2}{c}{\las}\\
      & orig & $\Delta$ & orig & $\Delta$ & orig & $\Delta$ \\    \midrule
    ar   &            60.6 &              2.6 &      38.9 &        3.1 &      25.6 &        2.6 \\
    de   &            64.6 &              6.0 &      39.6 &        5.2 &      29.5 &        6.2 \\
    en   &            58.4 &              3.0 &      37.8 &        2.9 &      25.7 &        3.3 \\
    fr   &            63.7 &              2.1 &      39.8 &        3.1 &      27.3 &        2.3 \\
    ru   &            58.4 &              2.0 &      36.1 &        2.3 &      22.9 &        2.0 \\
    \bottomrule
    \end{tabular}
\caption{$M_{R}$. Mid layers of mBERT architecture with completely random parameters}\label{tab:bl-randomrandom}
\end{table}

\begin{table}\centering
\begin{tabular}{lrrr}
\toprule
{} &  orig.  &  \ourdata  & $\Delta$ mean \\
\midrule
ar & 84.6 $\pm$ 0.13 & 78.9 $\pm$ 0.13 & 5.7 \\
de & 92.5 $\pm$ 0.04 & 90.8 $\pm$ 0.05 & 1.7 \\
en & 86.5 $\pm$ 0.10 & 84.8 $\pm$ 0.10 & 1.7 \\
fr & 89.9 $\pm$ 0.13 & 87.9 $\pm$ 0.25 & 2.0 \\
ru & 88.8 $\pm$ 0.00 & 87.3 $\pm$ 0.04 & 1.5 \\
\bottomrule
\end{tabular}
\caption{Mean and standard deviation \relacc of probing with different random seeds.}
\end{table}

\begin{table}\centering
\begin{tabular}{lrrrrrr}
\toprule
{} &  orig &  nonce  & $\Delta$ mean \\
\midrule
ar & 63.5 $\pm$ 0.18 & 56.0 $\pm$ 0.18 & 7.5 \\
de & 83.3 $\pm$ 0.08 & 80.0 $\pm$ 0.09 & 3.3 \\
en & 73.0 $\pm$ 0.31 & 69.3 $\pm$ 0.35 & 3.6 \\
fr & 75.5 $\pm$ 0.17 & 71.3 $\pm$ 0.33 & 4.2 \\
ru & 74.8 $\pm$ 0.02 & 71.6 $\pm$ 0.15 & 3.2 \\
\bottomrule
\end{tabular}
\caption{Mean and standard deviation \uas of probing with different random seeds.}
\end{table}

\begin{table}
\begin{tabular}{lrrrrrr}
\toprule
{} &  orig. &  nonce &  $\Delta$ means \\
\midrule
ar & 56.9 $\pm$ 0.13 & 48.0 $\pm$ 0.17 & 8.9 \\
de & 79.5 $\pm$ 0.06 & 75.6 $\pm$ 0.07 & 4.0 \\
en & 67.3 $\pm$ 0.29 & 63.6 $\pm$ 0.31 & 3.8 \\
fr & 70.8 $\pm$ 0.18 & 66.5 $\pm$ 0.32 & 4.3 \\
ru & 69.3 $\pm$ 0.01 & 66.2 $\pm$ 0.17 & 3.2 \\
\bottomrule
\end{tabular}
\caption{Mean and standard deviation \las of probing with different random seeds.}
\end{table}

\subsection{Random seeds}\label{app:exp-random-seeds}
The experimental setup introduced two random components: 
The probe's initial weights are randomly initialized, and the nonce data is created by randomly sampling words.  
To test the effect of these random components, we additionally run experiments with varying random seeds for each of these components in a pilot study. 
Per language, we initialize probes with 3 different random seeds, and each of these probes is evaluated on 5 different \ourdata datasets.
This means that per language, 3 predictions for original data and 15 predictions for nonce data are available. For all 3 evaluation metrics, the resulting scores across predictions are stable. Concretely, the largest observed standard deviation across predictions was 0.35 points \uas for English \ourdata data. 
Most of the other settings show much lower standard deviations. 
We argue that these random components do not have a large effect on the results, and that it is therefore sufficient to run the above probing experiments as described.

\subsection{Ignoring dependency relations in nonce data creation}\label{app:exp-nodeps} 
To compare our approach against the common nonce data creation practice of using POS tags, we test on a version of English and German nonce data that ignores dependency-specific information and only modifies the data based on POS tags and language-specific rules. %
The results are shown in Table~\ref{tab:pos-only-results}. 
The performance difference between the two versions of nonce data is larger for English than for German. 
This supports the fact that in English, dependency information is more easily extracted from positional information than in German: 
In the English data, the position of a word is a stronger cue for its dependency relation than in German.

\begin{table}\centering
  \scalebox{.8}{
  \begin{tabular}{l|rrrrrr}
  \toprule
  & \multicolumn{2}{c}{\relacc} & \multicolumn{2}{c}{\uas} & \multicolumn{2}{c}{\las}\\
  {} &  \ourdata & POS &  \ourdata & POS &  \ourdata & POS \\
  \midrule
  de & 91.0 & 90.1 & 81.0 & 79.7 & 76.2 & 75.0 \\
  en & 84.8 & 81.7 & 70.4 & 66.0 & 63.8 & 59.2 \\
  \bottomrule
  \end{tabular}
  }
  \caption{Results for English and German test sets with (\ourdataNoSpace ) and without (only POS) dependency information in the nonce data creation.}
  \label{tab:pos-only-results}
  \end{table}

\subsection{Layer dymanics}\label{app:layer-dynamics}

We train DepProbe on English data on all layers of mBERT and mGPT. 
On the one hand, this serves the purpose of selecting the layer used in all other experiments. 
\citet{muller-eberstein-etal-2022-probing} conducted the same experiment on mBERT to identify the layer that performs best on the probing task. 
The best layer for each task (6 for relation labeling, 7 for attachment) is then used in all other experiments. 
We repeat this process on mGPT to identify the highest performing layer for this model. 
In the case of mGPT, layer 12 performs best on both tasks. 
For selecting the best layer, we use the original test data (following existing work). 

On the other hand, the layerwise probing results highlight how syntactic and semantic information influence probing performance across layers. 
For \relaccNoSpace, mGPT's performance on both test sets develops relatively stable, with a sharp increase in the lowest layers and a slight decrease in the higher layers. 
The $\Delta$ to the \ourdata test set also increases during the lower layers and then stays quite stable. 
For mBERT, the pattern is similar, with a steeper decline in performance on the highest layer. 
For \uas (and \lasNoSpace), mGPT shows smaller changes across layers than for \relaccNoSpace. 
While the pattern appears jumpy, no individual difference between consecutive layers is larger than 3.5 points and the $\Delta$ to the \ourdata test set is stable. 
mBERT shows a clearer but steeper trend with best performance in the mid layers and worse performance in the higher and lower layers.

\begin{figure}
\includegraphics[width=.5\textwidth]{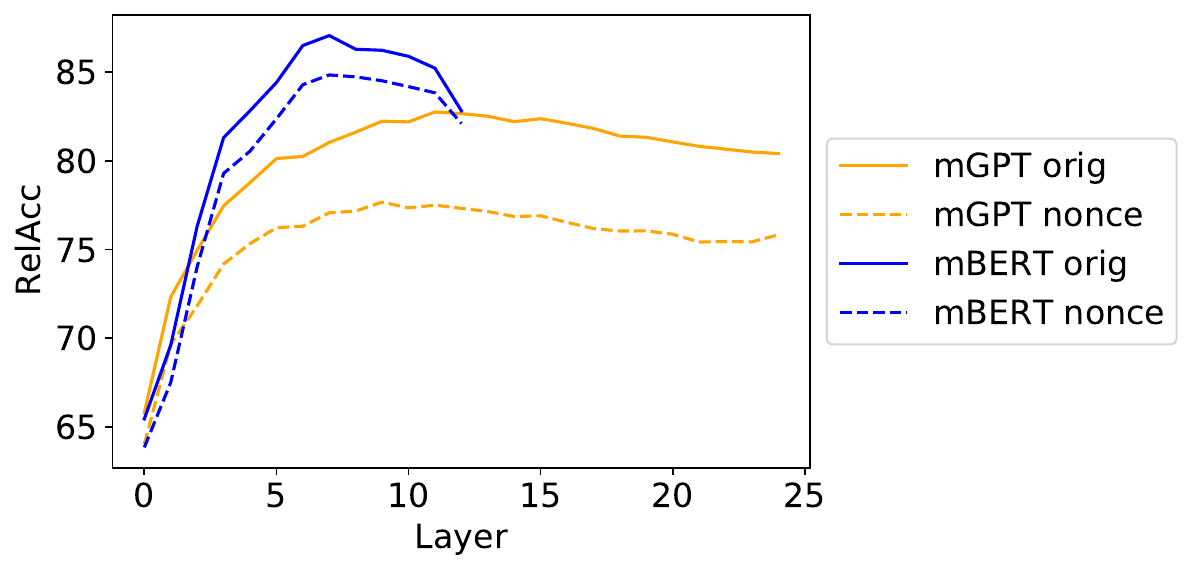}
\includegraphics[width=.5\textwidth]{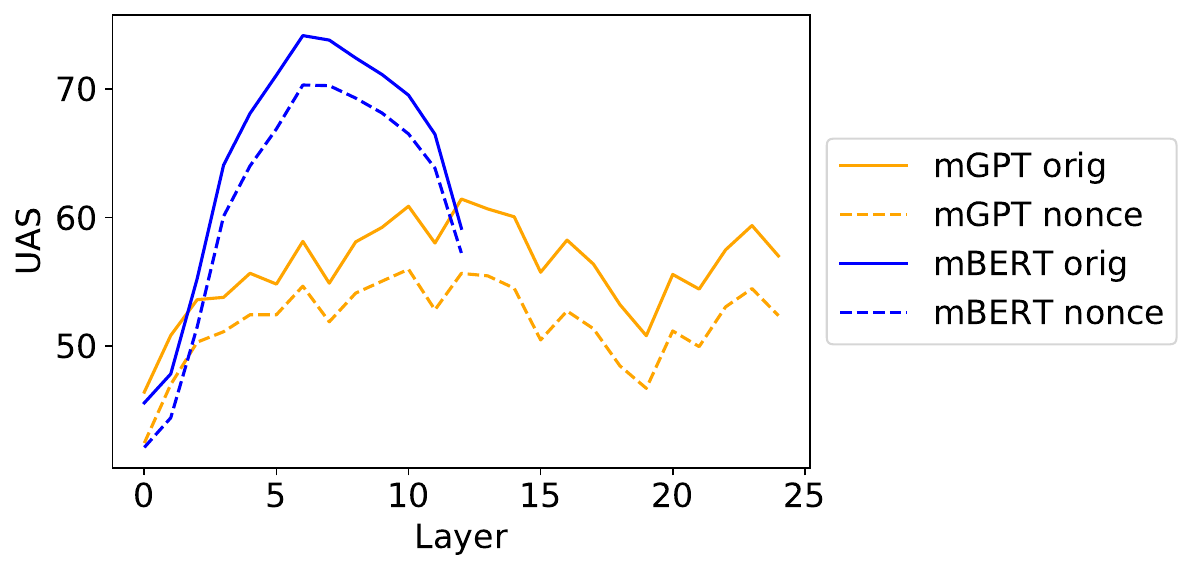}
\includegraphics[width=.5\textwidth]{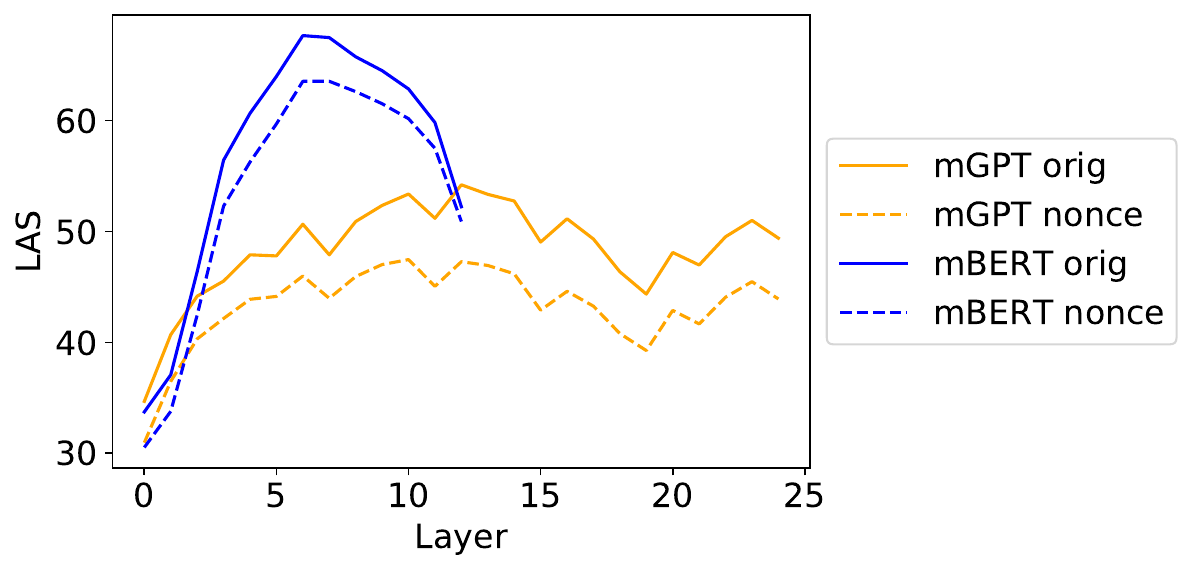}

\caption{Layerwise DepProbe performance for mBERT and mGPT.}
\end{figure}

\subsection{Probe Performance by edge direction}

\begin{table}
\centering
\begin{tabular}{lrr}
\toprule
{} &  left &  right \\
\midrule
ar & 0.67 & 0.31 \\
de & 0.32 & 0.63 \\
en & 0.37 & 0.56 \\
fr & 0.40 & 0.56 \\
ru & 0.42 & 0.53 \\
\bottomrule
\end{tabular}
\caption{Fraction of edges to the left and right in the respective UD test sets. Both columns do not sum to 1 because the root is not counted.}
\label{tab:edge-fraction}
\end{table}

\begin{table}\centering
    \scalebox{0.8}{
    \begin{tabular}{ll|rrrrrr}
        \toprule
        {} & & \multicolumn{2}{c}{\relacc} & \multicolumn{2}{c}{\uas} & \multicolumn{2}{c}{\las} \\
        {} & & l &  r &  l &  r &  l &  r \\
        \midrule
        \multirow{5}{*}{\rotatebox[origin=c]{90}{mGPT, orig.}}   &     ar &            81.8 &             91.9 &      60.0 &       56.0 &      52.0 &       53.5 \\
        & de &            88.3 &             94.5 &      77.0 &       86.7 &      71.1 &       82.9 \\
        & en &            75.4 &             87.4 &      54.3 &       58.7 &      44.5 &       53.3 \\
        & fr &            81.1 &             92.1 &      66.5 &       70.5 &      58.2 &       66.4 \\
        & ru &            84.9 &             91.5 &      71.7 &       77.0 &      64.1 &       71.8 \\\hline
        \multirow{5}{*}{\rotatebox[origin=c]{90}{mGPT, $\Delta$}} & ar &            10.3 &              2.3 &      15.6 &        6.6 &      16.9 &        6.7 \\
        & de &             5.8 &              2.5 &       9.3 &        5.9 &      10.9 &        7.0 \\
        & en &             7.0 &              3.9 &       5.5 &        5.4 &       6.6 &        6.4 \\
        & fr &             2.7 &              2.9 &       7.4 &        5.6 &       7.0 &        6.9 \\
        & ru &             4.2 &              2.1 &       7.1 &        4.1 &       7.7 &        4.8 \\\hline\hline
        \multirow{5}{*}{\rotatebox[origin=c]{90}{mBERT, orig.}} & ar &            77.8 &             92.7 &      63.1 &       60.9 &      52.7 &       58.8 \\
        & de &            85.7 &             95.9 &      75.4 &       87.8 &      67.7 &       85.1 \\
        & en &            75.4 &             94.5 &      65.8 &       78.0 &      52.6 &       75.3 \\
        & fr &            79.3 &             96.2 &      69.6 &       79.3 &      58.6 &       77.2 \\
        & ru &            81.4 &             93.3 &      70.5 &       77.7 &      60.3 &       74.0 \\\hline
        \multirow{5}{*}{\rotatebox[origin=c]{90}{mBERT, $\Delta$}} & ar &             7.6 &              1.1 &       8.8 &        4.3 &      10.4 &        4.3 \\
        & de &             2.6 &              1.2 &       4.3 &        3.1 &       5.1 &        3.7 \\
        & en &             2.9 &              1.8 &       3.0 &        4.5 &       3.1 &        4.9 \\
        & fr &             2.8 &              1.2 &       4.4 &        4.3 &       3.9 &        4.5 \\
        & ru &             1.9 &              1.1 &       3.6 &        3.0 &       3.2 &        3.0 \\
        \bottomrule
        \end{tabular}
    }
\caption{Performance for edges to the left and right}
\label{tab:edge-direction-results}
\end{table}

To further elaborate on the differences between decoded trees from ALMs and MLMs, 
we investigate the performance of DepProbe in our main experiment on edges in different directions. 
For all languages, edges to the left are defined as edges where the dependent has a lower index than the head (vice versa for edges to the right). 
The relative frequency of edges to the left and right is displayed in Tab.~\ref{tab:edge-fraction}. 
For all Indo-European (IE) languages, the majority of edges are to the right (left for Arabic).  
The results (Tab.~\ref{tab:edge-direction-results}) show that for the IE languages, edges to the right show higher performance than edges to the left. 
For ALMs, this is especially intuitive: The edges to the right are the ones where both head and dependent are in the context of the token on which the prediction is made. 
When considering only original data and IE languages, the performance difference between ALMs and MLMs interestingly shows that MLMs are better at predicting edges to the right, but not necessarily better at predicting edges to the left. 
Arabic consitutes a special case because the distribution of edges to the left and right is swapped. 
For \relacc and \ourdataNoSpace, edges to the left show lower performance than edges to the right, following the pattern in the other languages. 
For attachment, the more frequent edges to the left show higher performance than edges to the right (but only in original data). 
The performance drop on nonce data and \relacc is larger for edges to the left in all languages and both models. 
For attachment however, the pattern is more mixed. 
Since the performance drop on nonce data is secondary to the overarching research question of this particular experiment, we do not investigate this further. 

\section{Computational requirements}
All experiments were run on a single NVIDIA GeForce RTX 3090 GPU with 24GB of memory, and powered by renewable energy sources. 
Collecting the (pseudo-)perplexity scores for one combination of language, model, and scoring function takes between 5 minutes and 2 hours. 
The duration depends mostly on the dataset size and model architecture. Computing both versions of pseudo-perplexity for a single sentence requires $n$ forward passes for a sentence of length $n$, while $PPL$ computation requires just one forward pass. 
Training a single DepProbe model takes between 10 minutes and 3 hours, depending on the size of the LM and treebank. Probe inference is faster. 
From these values, we estimate that the complete set of experiments reported in this paper can be conducted with our hardware in 25-30 hours. 
Creating SPUD treebanks does not require GPU access and takes several minutes on a CPU and 64GB RAM.

\end{document}